\documentclass[sn-mathphys-num]{sn-jnl}

\usepackage{physics}%
\usepackage{graphicx}%
\usepackage{multirow}%
\usepackage{amsmath,amssymb,amsfonts}%
\usepackage{amsthm}%
\usepackage{mathrsfs}%
\usepackage[title]{appendix}%
\usepackage{xcolor}%
\usepackage{caption}%
\usepackage{textcomp}%
\usepackage{manyfoot}%
\usepackage[utf8]{inputenc}
\usepackage{booktabs}%
\usepackage{algorithm}%
\usepackage{algorithmicx}%
\usepackage{algpseudocode}%
\usepackage{listings}%
\usepackage{subcaption}%
\usepackage{array}%
\usepackage{ragged2e}%

\theoremstyle{thmstyleone}%
\theoremstyle{thmstyletwo}%

\theoremstyle{thmstylethree}%

\begin{document}

\title[Article Title]{Quantum Gradient-Based Approach for Edge and Corner Detection Using Sobel Kernels}


\author[1]{\fnm{Mohammad Aamir} \sur{Sohail}}\email{mdaamir@umich.edu}

\author[2]{\fnm{Gabriela} \sur{Pinheiro}}\email{gabrielapc@id.uff.br}

\author*[3]{\fnm{Yasemin} \sur{Poyraz Koçak}}\email{yasemin.poyraz@iuc.edu.tr}

\author[4]{\fnm{Batuhan} \sur{Hangün}}\email{batuhanhangun@gmail.com}

\author[5]{\fnm{Emre} \sur{Camkerten}}\email{a12325613@unet.univie.ac.at}

\author[3]{\fnm{Simge} \sur{Yiğit}}\email{simge.yigit@ogr.iuc.edu.tr}

\author[3]{\fnm{Hafize Asude} \sur{Ertan}}\email{hafizeasudeertan@ogr.iuc.edu.tr}

\affil[1]{\orgdiv{Department of Electrical Engineering and Computer Science}, \orgname{University of Michigan}, \orgaddress{\street{1301 Beal Avenue}, \city{Ann Arbor}, \postcode{48109-2121}, \state{Michigan}, \country{USA}}}

\affil[2]{\orgdiv{Department of Computer Science}, \orgname{Universidade Federal Fluminense}, \orgaddress{\street{Campus da Praia Vermelha}, \city{Niterói}, \postcode{24210-346}, \state{Rio de Janeiro}, \country{Brazil}}}

\affil*[3]{\orgdiv{Department of Computer Technologies}, \orgname{İstanbul University-Cerrahpaşa}, \orgaddress{\street{Buyukcekmece}, \city{İstanbul}, \postcode{34500}, \country{Turkiye}}}

\affil[4]{\orgdiv{Department of Computer Science}, \orgname{University of Vienna}, \orgaddress{\street{Währinger Straße 29}, \city{Vienna}, \postcode{1090}, \country{Austria}}}

\affil[5]{\orgdiv{Department of Computer Engineering}, \orgname{Yıldız Technical University}, \orgaddress{\street{Esenler}, \city{İstanbul}, \postcode{34220}, \country{Turkiye}}}


\abstract{Edge detection refers to the process of identifying points in a digital image where the intensity changes sharply, typically indicating object boundaries or structural features. Corners are locations where the gray value intensity changes abruptly in multiple directions and are widely used in feature extraction, object tracking, and 3D modeling. In this study, we present a quantum implementation of Sobel-based edge detection and Harris-style corner detection. Two quantum image encoding methods—Flexible Representation of Quantum Images (FRQI) and Quantum Probability Image Encoding (QPIE)—are employed to encode the input data, and their behavior is comparatively analyzed. The proposed approach introduces a quantum gradient computation scheme based on lag-2 differences, which enables the evaluation of gradient-like features in superposition. To improve detection quality and eliminate false positives, a classical post-processing step is applied to candidate corner points identified by the quantum circuit. The results demonstrate that the proposed quantum circuits produce edge and corner detection outputs that are \textbf{qualitatively} consistent with classical Sobel and Harris operators. Furthermore, it is observed that the QPIE-based configuration yields \textbf{more stable and structurally coherent results} compared to its FRQI-based counterpart, particularly under limited measurement shots. While the gradient computation can be performed efficiently at \textbf{the circuit subroutine level}, the overall computational cost remains dominated by state preparation, measurement, and classical post-processing steps. \textbf{All experiments are conducted under noiseless simulation conditions, and practical performance on NISQ hardware may be affected by noise, measurement overhead, and finite-shot statistics.} Therefore, the present work primarily demonstrates a functional and scalable quantum realization of classical edge and corner detection methods, rather than an end-to-end computational speedup.}




\maketitle

\section{Introduction}\label{sec1}

Digital image processing refers to processing digital images such as acquiring an image of the specific area, pre-processing that image, extracting or segmenting the individual characters, describing the characters in a form suitable for computer processing, and recognizing those individual characters by means of algorithms on a digital computer \cite{gonzalez2008digital}. Among the key operations in digital image processing are edge and corner detection, both of which have an important role on describing and extracting characters \cite{ref2}. \\
Edge detection algorithms determine sharp discontinuities in an image, which typically correspond to object boundaries. It plays a vital role in various domains such as medical imaging and autonomous driving. Classical edge detectors such as Sobel\cite{vincent2009descriptive}, Prewitt\cite{yang2011improved}, and Canny\cite{canny} focus on detecting gradients in intensity to reveal meaningful structural contours. The edge detection output often serves as input for further processing stages, such as corner detection\cite{gonzalez2008digital}.
Corners are locations where the gray value intensity changes suddenly in two or more directions. Corner detection algorithms are especially used for detecting moving objects, 3D modeling \cite{xu20223d}, image registration \cite{misra2022feature}, image mosaicing \cite{dadhoreimage}, panorama stitching \cite{abbadi2021review} and object tracking through graphical matching techniques \cite{huang2024survey}, \cite{schmid2000evaluation}.\\

Quantum computing has emerged as a promising tool overcoming the limitations of classical algorithms. Using quantum phenomena such as superposition and entanglement, quantum algorithms can perform certain computations exponentially faster than their classical counterparts. Quantum Image Processing (QIP) is an emerging interdisciplinary field at the intersection of quantum computing and digital image processing, and it has garnered increasing attention with the rise of quantum machine learning applications. Recent comprehensive surveys have highlighted both the significant progress and remaining challenges in this field \cite{yan2025lessons, chakraborty2022challenges, dhar2025systematic}, emphasizing the need for more efficient quantum algorithms and practical implementations on noisy intermediate-scale quantum (NISQ) devices.

The development of quantum methods has led to substantial improvements in computational efficiency, noise resilience, and scalability. In \cite{ref7}, with the purpose of reducing the required number of qubits, probability amplitudes and computational basis states to encode pixel values and positions have been used respectively. Also for edge detection, a quantum algorithm that completes the task with only one single-qubit operation, independent of the size of the image have been proposed. 
In \cite{ref8}, using the idea of classical filtering and Tacchino's quantum machine learning algorithm, a hybrid method is introduced for quantum edge detection. In this method, two filter masks highlighting vertical and horizontal edges are used in combination. In \cite{ref2}, a new quantum fast corner detection algorithm was proposed by leveraging the advantage of quantum parallelism. Although the algorithm's time complexity and quantum delay do not increase with the growth in image size and number, and its time complexity is significantly lower than that of the classical fast corner detection algorithm, the number of gates used is quite high due to the algorithm's numerous computationally intensive parts. The development of hybrid quantum-classical pipelines represents a growing trend in quantum image processing, combining the advantages of quantum parallelism with classical post-processing techniques to achieve practical implementations on current quantum hardware \cite{deb2024compression, ebrahimpour2024survey, wang2024haar}.\\
The Harris corner detector, introduced in \cite{ref9}, remains one of the most influential methods for corner detection in image processing. The method identifies corners by computing the eigenvalues of the second-moment matrix, or structure tensor, of the intensity gradient in a local neighborhood of the image. A point in the image is classified as a corner if the eigenvalues are both significant, indicating that intensity changes in multiple directions. Despite its success, the Harris detector faces limitations, particularly in real-time and large-scale applications due to its computational complexity and sensitivity to noise. Recent improvements to the Harris detector have incorporated fractional-order techniques \cite{lavin2024fractional} and multi-scale analysis approaches \cite{ref3}, while quantum implementations offer the potential for circuit-level parallelism through simultaneous gradient computation.\\
\noindent In this study, we propose the Quantum Harris Corner Detection algorithm. It aims to address the computational related challenges of the original Harris detector. To compare the success rate of Quantum Harris Corner Detection algorithm, different image encoding methods such as Flexible Representation of Quantum Images (FRQI) \cite{ref10} and Quantum Probability Image Encoding (QPIE)\cite{ref7} methods are implemented. The dataset used in this study is obtained from the publicly available work in \cite{soria2023teed}. In addition to corner detection, we also integrate a Quantum Hadamard Edge Detection (QHED) algorithm to identify edge structures using a quantum-enhanced approach. The QHED model performs lag-2 differencing followed by Hadamard-based interference to emphasize high-frequency changes.
To compute both edge and corner features efficiently, we design a Quantum Gradient Kernel Circuit, which performs lag-2 differencing via quantum permutation operators. This is implemented separately for both FRQI and QPIE representations.
Following quantum-based gradient estimation, classical Sobel-based post-processing is applied to further refine the detected features. This hybrid method enables improved detection performance and better localization.
Through extensive experiments and analysis, the following contributions of this study are demonstrated:
\begin{itemize}

\item Implementation of a quantum gradient kernel circuit to compute lag-2 differences as a core operation for edge and corner detection.

\item Design and comparison of QHED and QHCD algorithms under both QPIE and FRQI image encoding frameworks.

\item Introduction of a novel quantum-classical hybrid pipeline that combines quantum amplitude-based differencing with classical Sobel post-processing.

\item Demonstration of corner detection exploiting quantum parallelism, where gradient information for all pixel positions is encoded simultaneously within a single quantum state, enabling efficient gradient computation via amplitude-based interference.

\item Efficient derivation of $I_x$ and $I_y$ with minimal circuit depth, reducing quantum hardware resource requirements.

\item A promising application framework for Quantum Image Processing, paving the way for further research in quantum-enhanced feature extraction and image understanding.
 
\end{itemize}

\noindent Because of these advantages, the newly proposed algorithm is more flexible and better suited to be the fundamental model for quantum image processing than all other existing quantum corner detection algorithms.\\
The rest of the paper is organized as follows: In Section 2, the materials and methods are presented. This includes the quantum image representation models (QPIE and FRQI), the image retrieval process, and the design of the Quantum Gradient Kernel Circuit for computing lag-2 differences. Section 2 also covers the edge and corner detection frameworks, including both classical (Sobel, Harris) and quantum-based (QHED, QHCD) implementations. Section 3 presents the experimental setup and results, with a comparative evaluation of edge and corner detection performance. Finally, Section 4 provides the conclusion and outlines future research directions. 

\subsection{Related Work}

In literature, various algorithm have been developed to detect edges and corners. \cite{feng2025multiscale} proposes a novel color edge detection method combining collaborative filtering (CBM3D) with multiscale gradient fusion to enhance edge quality and noise robustness. The method demonstrates superior performance over traditional techniques such as Color Sobel, SE, and AGDD, particularly in terms of PR curve, AUC, and FOM metrics. In \cite{zangana2024advancements}, classical and modern edge detection techniques have been evaluated, advancements involving deep learning, fuzzy logic, and optimization algorithms have been highlighted by emphasizing their applications in fields such as medical imaging and autonomous systems. The study also identifies current limitations and outlines future research directions to enhance edge detection accuracy and robustness. Existing curvature scale-space (CSS) methods used to detect corner points in images with complex details may cause miss-detected true corners or detected false corners \cite{hennig2021box}. In  \cite{ref3}, a novel CSS-based corner detector is proposed by incorporating mathematically derived scale-space properties of planar curves and corner points into the developed trajectory tracing algorithm, called the scale-space behavior-guided trajectory tracing (SBTT) and the proposed algorithm showed better results than other nine state-of-the-art methods. In \cite{ref4}, a composite model of intensity, pattern, curvature, and scale is proposed. In this model, intensity, pattern, and curvature differences are all formulated by considering 8-neighboring pixel blocks. Secondly, some scale-based global scale importance factors are formulated based on the contour distribution and corner distribution. Finally, a high-performance corner detector (IPCS) is derived based on the corner measure and the importance factors.\\ 
Existing deep learning methods for corner detection focus only on corner points with high repeatability but neglect the importance of corner point features. To overcome this limitation, the authors of \cite{ref5} introduced a new corner detection architecture and designed a network capable of efficiently learning corner feature information from images. The proposed model achieved significantly superior performance compared to state-of-the-art methods.\\

The first step of quantum image processing is the integration of classical image data into quantum circuits, a process known as quantum image encoding. Several prominent quantum image encoding methods have been proposed in the recent literature. Among these, the Flexible Representation of Quantum Images (FRQI) \cite{geng2023improved}, Novel Enhanced Quantum Representation (NEQR) \cite{NEQR}, Quantum Boolean Image Processing (QBIP) \cite{mastriani2015quantum}, and Quantum Probability Image Encoding (QPIE) \cite{ref7} stand out as foundational models. In addition, Multichannel Representation for Quantum Images (MCQI) \cite{sun2013rgb}, Color Quantum Image Representation (CQIR) \cite{caraiman2012image}, Quantum Amplitude Encoding of Color Pixels (QUALPI) \cite{zhang2013novel}, State-based Quantum Representation (SQR) \cite{yuan2014sqr}, Quantum Superposition Multi-Channel (QSMC) $\&$ Quantum Superposition Normalized Color (QSNC) \cite{li2013image}, and Novel Adaptive Quantum State with Superposition (NAQSS) \cite{li2014multi} have been developed as extended or alternative frameworks. Recent developments have also introduced Quantum Polar Representation (QPR)\cite{zhang2013novel} and Real Ket Representation (RKR)\cite{zhou2023quantum} as novel approaches aiming to reduce quantum gate complexity and enhance image fidelity under noise. Furthermore, advances in quantum image representation include density matrix-based approaches for open quantum systems \cite{hu2024density}, qutrit-based representations for enhanced storage efficiency \cite{taheri2024qutrit}, and improved color image representations such as INCQI and AFQIRHSI that support multi-channel processing \cite{su2021incqi, alwan2025afqirhsi}.

\noindent In addition, researchers have begun to explore quantum versions of classical image processing techniques. In the context of quantum image processing has demonstrated potential in accelerating tasks such as image compression, filtering, and feature detection. In \cite{ref6} a comprehensive review of quantum image processing (QIP) algorithms, which include methods for quantum image representation, filtering, and feature detection, are provided. 
Recent quantum edge detection algorithms have achieved notable improvements, including methods based on the Laplacian of Gaussian operator \cite{yuan2024log}, algorithms optimized for noisy images \cite{chetia2024noisy}, and approaches utilizing the difference of Gaussian operator \cite{fan2023gaussian}.

In parallel, quantum methods have also been explored for image pre-processing tasks such as denoising, which is often a critical step before feature extraction. Recent works include a quantum deep convolutional generative adversarial network for medical image denoising \cite{nandal2025qdcgan}, and a quantum synthetic aperture radar (SAR) image denoising algorithm based on grayscale morphology that removes multiplicative (speckle) noise by processing all pixels simultaneously \cite{wang2024qsar}. These developments suggest that quantum image processing is maturing into a full pipeline covering acquisition, pre-processing, and feature extraction.

Beyond static images, the lag-based differencing concept central to our quantum gradient kernel extends naturally to the temporal domain, where inter-frame differences can be used to detect moving targets in video. Recent work has begun to explore this emerging area of quantum video processing: Liu et al.~\cite{liu2023qmts} proposed a quantum moving target segmentation algorithm for grayscale video that leverages three-frame differencing to simultaneously compute pixel-wise differences across all adjacent frames. Building on this direction, Wang et al.~\cite{wang2024qmts_bg} introduced a background-difference variant that simultaneously models the background across all frames using a quantum subtractor. These algorithms share the structural parallel with our work of exploiting quantum superposition to evaluate differencing operations across all positions in a single coherent execution, and collectively indicate that quantum-temporal analysis is a promising extension of quantum image processing.

\section{Material and Methods}\label{sec2}
In this section, two different quantum image encoding models (QPIE and FRQI) are explained for efficient image representation. These encoding models enable quantum-based image retrieval and serve as the foundation for subsequent operations. Then, Quantum Gradient Kernel Circuit is introduced, where lag-2 difference computations are performed using both QPIE and FRQI. Then classical post-processing techniques are applied to the result data to implement edge and corner detection framework using Sobel kernel. Finally, both classical and quantum approaches for edge and corner detection are presented, demonstrating how structural features are effectively identified in the quantum domain.
\subsection{Quantum Image Representation}\label{subsec1}
The first step in the quantum image processing framework involves encoding images into a quantum state. For this, we utilized two different methods for quantum image representation: Quantum Probability Image Encoding (QPIE) and Flexible Representation of Quantum Images (FRQI).
\subsubsection{Quantum Probability Image Encoding (QPIE)}\label{subsubsec1}
The QPIE method provides an easy way to represent an image using a quantum state proposed in \cite{ref7}. In this method, each pixel value of the image is encoded in the probability amplitudes, and the pixel positions are represented in the computational basis states of a quantum state. One advantage of this method is the low number of qubits needed to encode an image. For a 2D image of size $2^n \times 2^n$, it only requires $2n$ qubits.
Consider a rectangular 2D image \(I\) of dimensions \( m \times n \). This image can be represented as a {vector} as seen in Equation \ref{eq1} by concatenating its rows sequentially to form 
\begin{equation}
     I_{\text{vec}} = (I_{1,1}, I_{1,2}, \ldots, I_{1,n},I_{2,1}, \ldots, I_{m,n})
     \label{eq1}
\end{equation}

\noindent Once normalized, $I_{\text{vec}}$ serves as the probability amplitudes of a quantum state that encodes the entire image. In this representation, the computational basis state \( \vert k \rangle \) encodes the position of each pixel $(i,j)$, and the corresponding coefficient encodes the pixel intensity value. Mathematically, the encoded image quantum state \( \vert I \rangle \) is defined by Equation \ref{eq2}:

\begin{equation}
    \vert I \rangle = \sum_{i=0}^{2^r - 1} c_i \vert i \rangle, \quad r = \lceil \log_2(mn) \rceil
    \label{eq2}
\end{equation}

where $c_i = I_{\text{vec}}(i)/\norm{I_{\text{vec}}}_2$ is the normalized pixel value of the $i$-th element of $I_{\text{vec}}$ and $r$ is the number of qubits required to encode the image. Since the pixel intensities are encoded as normalized probability amplitudes, any gradient quantity $\Delta c_i$ obtained from the quantum gradient kernel can be rescaled to the original intensity domain using Equation \ref{gradient_rescaling}:
\begin{equation}
\Delta I_i = \| I_{\mathrm{vec}} \|_2 \, \Delta c_i,
\label{gradient_rescaling}
\end{equation}
where $\| I_{\mathrm{vec}} \|_2$ is the global normalization factor retained during encoding. This rescaling ensures that the Sobel-based gradient magnitude and the corner response $R(M)$ are evaluated using absolute intensity gradients rather than relative ones.


\subsubsection{Flexible Representation of Quantum Images (FRQI)}\label{subsubsec2}
FRQI has been developed to efficiently encode pixel and position information within images into a quantum state \cite{ref10}. In this method, each pixel value of the image is encoded into angles associated with an ancilla qubit, and the rest of the qubits encode the pixel's position. This method also requires a small number of qubits to encode an image, having only an additional ancilla qubit compared to QPIE. For an image size of $2^n \times 2^n$, it requires $(2n + 1)$ qubits. 
Consider a rectangular 2D image  $I$  of dimensions $ m \times n$ . Similar to QPIE, in FRQI, the image is first converted into a vector by concatenating its rows sequentially like Equation (\ref{eq3}):
\begin{equation}
I_{\text{vec}} = (I_{1,1}, I_{1,2}, \ldots, I_{1,n}, I_{2,1}, \ldots, I_{m,n}).
\label{eq3} 
\end{equation}

\noindent For $b$-bit grayscale image, each element of \( I_{\text{vec}}(i) \in [0, 2^b - 1] \) is then mapped to a rotation angle \( \theta_i \in [0, \pi/2] \) by using Equation \ref{eqn:frqi_enc1} and \ref{eqn:frqi_enc2}:

\noindent\text{1. Nonlinear encoding:}
\begin{equation}
    \theta_i = \arcsin\left( \frac{I_{\text{vec}}(i)}{2^b - 1} \right)
\label{eqn:frqi_enc1} 
\end{equation}
\text{2. Linear encoding:} 
\begin{equation}
    \theta_i = \left( \frac{I_{\text{vec}}(i)}{2^b - 1} \right)\cdot \frac{\pi}{2}
\label{eqn:frqi_enc2} 
\end{equation}\\
Finally, the quantum state of the encoded image \( |I_{e} \rangle\) can be expressed as Equation \ref{eq4}:
\begin{equation}
\vert I \rangle=\frac{1}{\sqrt{2^r}} \sum_{i=0}^{2^{r}-1}(\cos{{\theta}_i}\vert 0 \rangle+\sin{{\theta}_i}\vert 1 \rangle) \otimes \vert i \rangle
\label{eq4}
\end{equation}
where 
$r = \lceil \log_2(mn) \rceil$ is the number of qubits required to encode the pixel position. 
The choice between these two mappings depends on the downstream application. The nonlinear $\arcsin$-based encoding is particularly suitable for tasks such as image reconstruction, as it directly encodes the normalized pixel intensities into the amplitude of the quantum state.
In comparison, linear encoding distributes pixel values uniformly over the interval $[0, \pi/2]$, giving more focus to the relative intensity variation between pixels and being better suited for applications like edge and corner detection that are highly dependent on contrast. This method avoids the nonlinear distortion introduced by the arcsin function, thereby preserving relative intensity differences more accurately.\\
\noindent One advantage of FRQI is that it can be used to encode a $b$-bit colored pixel with $(R,G,B)$ values. For example, consider $i$-th pixel with $(R_i,G_i,B_i)$ value, the encoded angle is expressed as:

\begin{equation}
{\theta}_i := \arcsin\left(\frac{R_i}{(2^b\!-\!1)}+\frac{G_i}{(2^b\!-\!1)^2}+\frac{B_i}{(2^b\!-\!1)^3}\right)
\label{eq5}
\end{equation}

\noindent After retrieving the angle $\theta_i$, the encoding process can be reversed to extract the original $(R,G,B)$ values. This is done by expressing $(2^b\!-\!1)^3\sin(\theta_i)$ as a base number $(2^b\!-\!1)$.

\noindent For example, two neighboring pixels with grayscale intensities $v_i$ and $v_j$. The angular difference governing the quantum gradient response can be written as
\begin{equation}
\Delta \theta = \theta(v_i) - \theta(v_j).
\label{eq:delta_theta}
\end{equation}
\noindent Under linear FRQI encoding, the mapping from pixel intensity to rotation angle is linear, implying that $\Delta \theta$ scales proportionally with the absolute intensity difference $|v_i - v_j|$. This results in a uniform sensitivity to intensity variations across the full grayscale range, which is essential for reliable gradient-based edge and corner detection.
\noindent In contrast, for the nonlinear arcsin-based encoding, the local sensitivity of the rotation angle to intensity variations is given by
\begin{equation}
\frac{d\theta}{dv} = \frac{1}{\sqrt{1 - \left( \frac{v}{2^b - 1} \right)^2 }},
\label{eq:arcsin_derivative}
\end{equation}
which exhibits a non-uniform behavior depending on the intensity level. Specifically, this nonlinear mapping compresses or amplifies intensity differences unevenly, leading to potential distortion of local gradient magnitudes. Such distortion can adversely affect the accuracy of edge and corner localization, particularly in regions with low contrast.
\noindent Therefore, linear encoding is more suitable for gradient-driven quantum image processing tasks, as it preserves relative intensity differences more faithfully and enables more stable amplitude-based gradient estimation.

\noindent Another advantage of FRQI is its simpler implementation, as pixel intensities are encoded using rotation angles. This mapping enables a quantum circuit design that utilizes a polynomial number of controlled rotation gates Theorem 1 in \cite{ref10}, \cite{geng2023improved,ref13,khan2019improved}. In contrast, QPIE is more challenging to implement because it encodes image data as a probability amplitude, complicating the circuit design and potentially not guaranteeing implementation with a polynomial number of gates in general.

\begin{figure}[!htb]
    \centering
\includegraphics[scale=0.26]{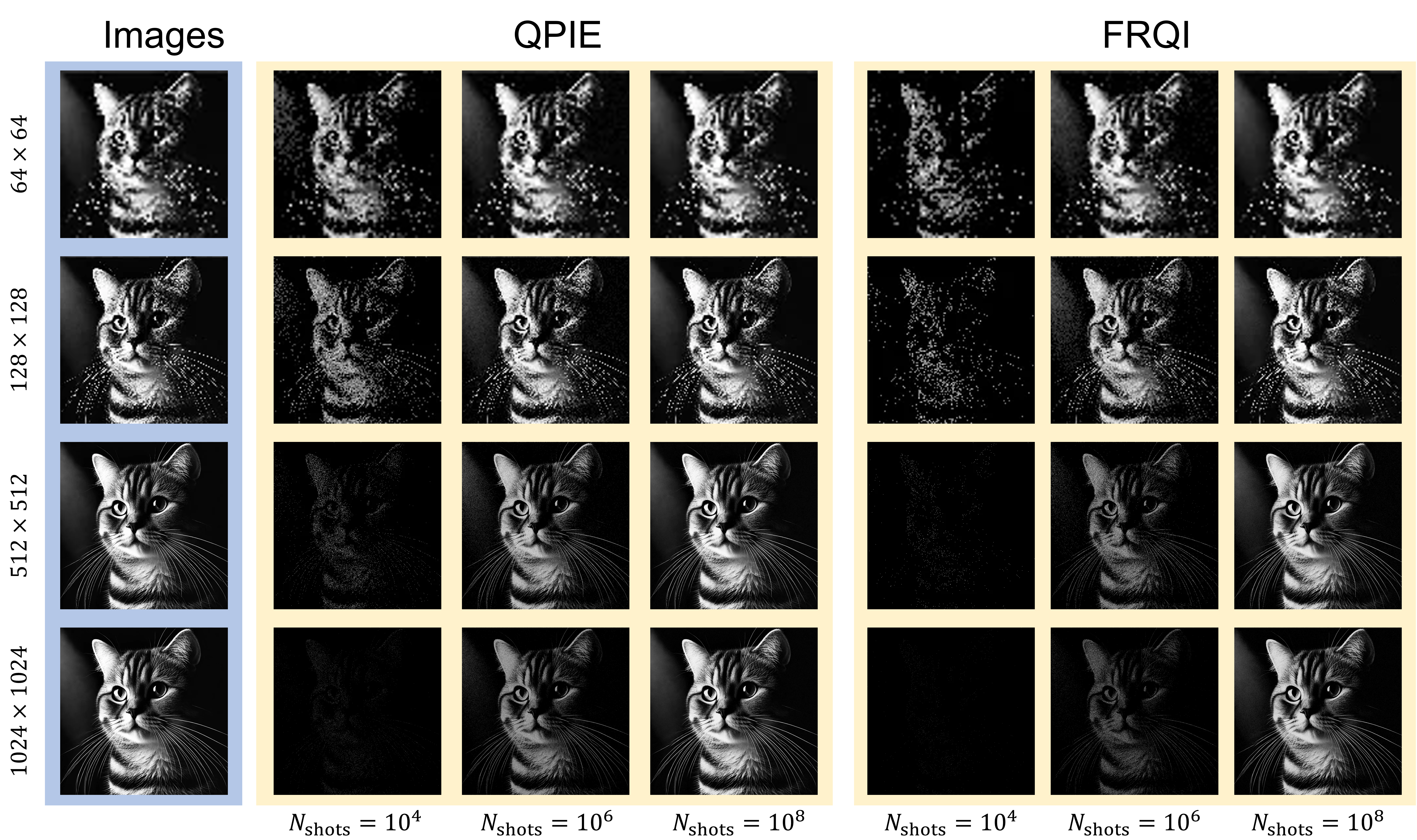}
    \caption{Comparison of QPIE and FRQI image encoding methods. The leftmost column shows grayscale images of varying resolutions: $64 \times 64$, $128 \times 128$, $512 \times 512$, and $1024 \times 1024$. The middle columns display the reconstructed images using the QPIE method for different numbers of shots ($N_{\text{shots}} = 10^4, 10^6, 10^8$). The rightmost columns show the reconstructed images using the FRQI method for the same shot counts. The QPIE method demonstrates higher fidelity in image reconstruction compared to FRQI, especially at lower shot counts.}
\label{fig:QPIE_FRQI_comparison}
\end{figure}

\subsection{Image Retrieval using QPIE and FRQI}\label{subsec2}

Once an image is encoded into a quantum state, measuring the quantum encoding circuit produces different basis states based on the encoding method. 
In the QPIE method, the retrieval process is straightforward: the encoded state is measured in the computational basis, yielding a state of the form $ \vert \texttt{bin}(i) \rangle$, where $i\in [0,2^{r}\!-\!1]$, \texttt{bin}($\cdot$) converts a decimal into a binary string, and each basis state corresponds directly to a pixel’s position and its associated probability amplitude encodes the normalized intensity. By applying the inverse of the normalization, the original pixel intensity is recovered. More formally, given the number of measurement shots $N$ and the number of times $i$-th state is observed $N_i$. The $i$-th amplitude of the encoded state can be approximated as Equation (\ref{eq6e}):
\begin{equation}
P(\text{observing the sequence }{`\!\!\texttt{ bin}(i)\text{'}}) = \frac{N_i}{N} = c^2_i
\label{eq6e}
\end{equation}

\noindent Finally, we get the $i$-th pixel value using Equation (\ref{eq7e})

\begin{equation}
{I_{\text{vec}}(i)} = c_i \norm{I_{\text{vec}}}_2 = \sqrt{\frac{N_i}{N}}\norm{I_{\text{vec}}}_2
\label{eq7e}
\end{equation}

\noindent In the FRQI method, assuming nonlinear $\arcsin$-based encoding, after the measurement in the computational basis, we get a state of the form $\ket{a}\ket{\texttt{bin}(i)}$, where $a\in \{0,1\}$ denotes the output of the ancilla qubit. Let $N_{1i}$ be the number of times $\ket{1}\ket{\texttt{bin}(i)}$ is observed. Then, the probability of measuring the state $\ket{1}\ket{\texttt{bin}(i)}$ is given as Equation (\ref{eq6})
\begin{equation}
    P(\text{observing the sequence }{`1} {\texttt{bin}(i)\text{'}}) \!= \frac{N_{1i}}{N} 
    \label{eq6}
\end{equation}

Therefore, the $i$-th pixel value can be expressed following Equation (\ref{eq7})
\begin{equation}
{I_{\text{vec}}(i)} = (2^b-1) \sqrt{2^r \frac{N_{1i}}{N}}
\label{eq7}
\end{equation}

\noindent To accurately recover pixel values from both QPIE and FRQI encoded states requires a number of measurements that scale exponentially with the number of qubits, particularly 
$O(2^r)$, where $r$ is the number of qubits required to encode an image. In Figure ~\ref{fig:QPIE_FRQI_comparison}, we compare image retrieval using QPIE and FRQI across four image sizes and varying numbers of measurement shots. The results show that achieving a close approximation to the original image at resolutions of \(64\times64\) and \(128\times128\) generally requires at least \(10^6\) shots, whereas larger size images demand around \(10^8\) shots. Moreover, for the same number of measurement shots, QPIE consistently yields more accurate reconstructed images than FRQI. This discrepancy arises from FRQI’s additional ancillary qubit, which necessitates more measurements to accurately approximate pixel values. 

\begin{figure}[!htb]
    \centering
    \includegraphics[width=0.9\textwidth]{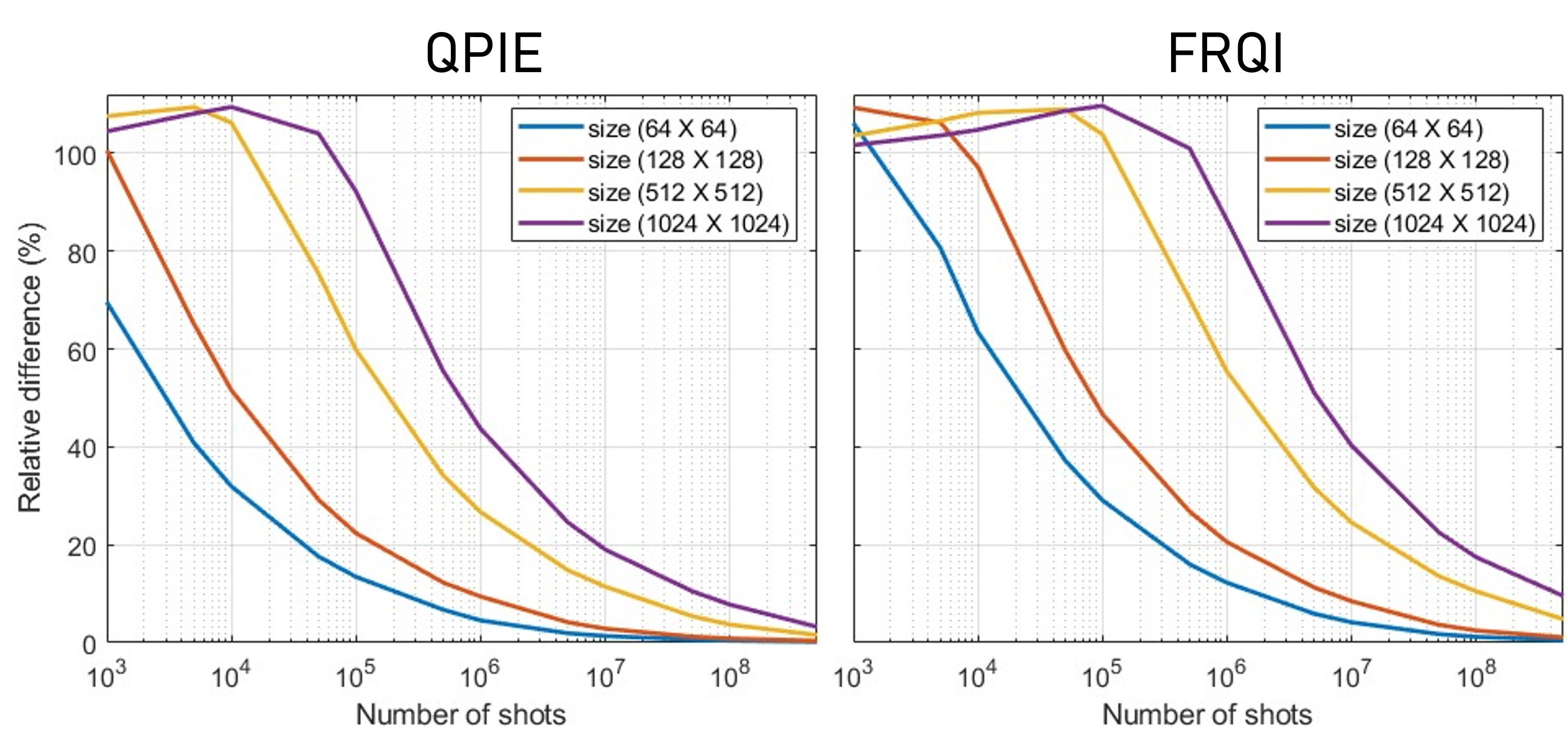}
    \caption{Relative difference between the original image and reconstructed image after QPIE and FRQI encoding.}
    \label{fig:relDiff_CAT}
\end{figure}
\begin{figure}[!htb]
    \centering
    \includegraphics[width=0.9\textwidth]{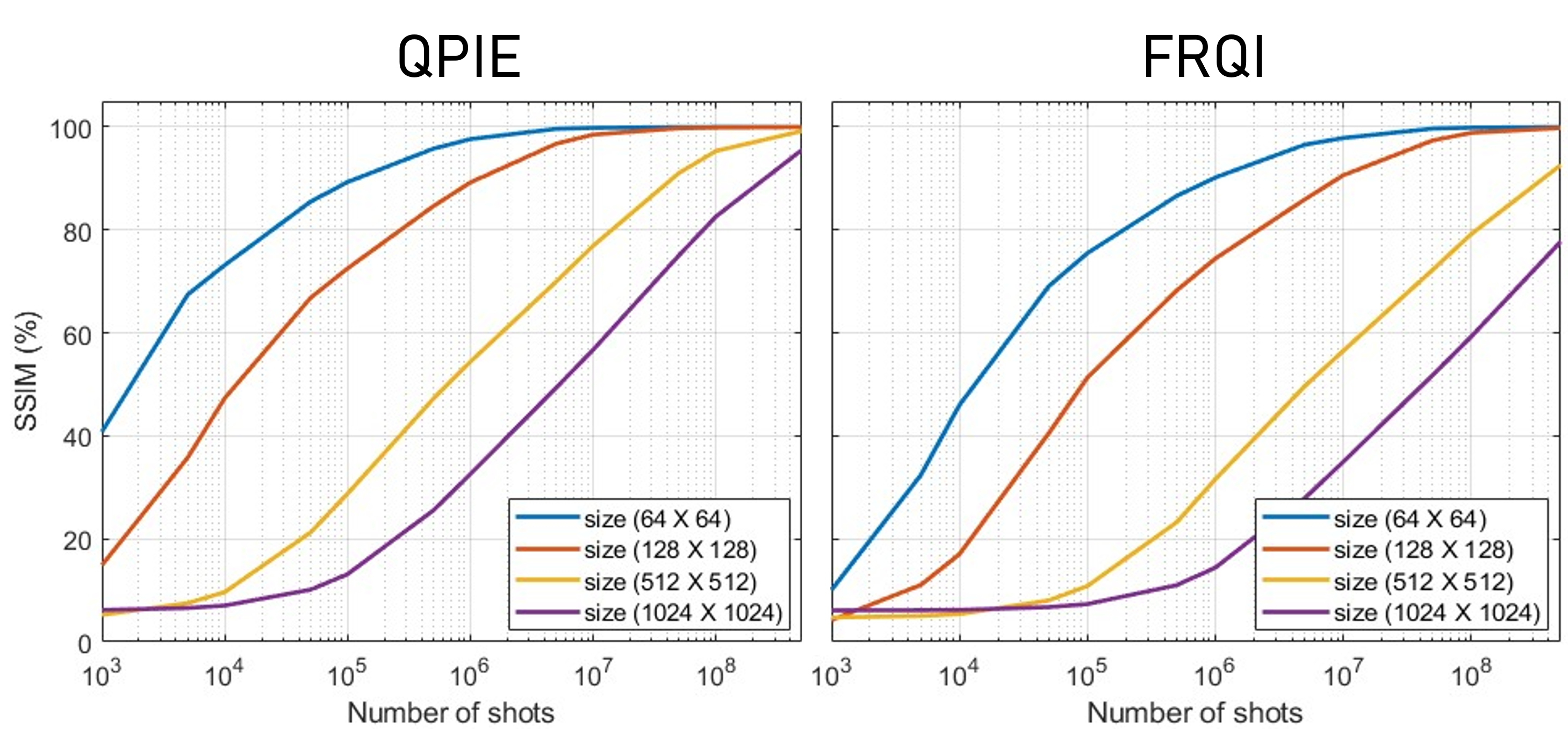}
    \caption{Structural similarity index measure (SSIM) between the original image and reconstructed image after QPIE and FRQI encoding.}
    \label{fig:ssim_CAT}
\end{figure}

\noindent Figures \ref{fig:relDiff_CAT} and \ref{fig:ssim_CAT} further 
illustrate these findings: plots of the relative difference between original and reconstructed images, as well as the structural similarity index measure (SSIM), demonstrates that FRQI exhibits both higher relative difference and lower SSIM compared to QPIE for the same number of measurement shots, especially for larger images.
\subsection{Quantum Gradient Kernel Circuit}\label{subsec3}
Gradient-based techniques are fundamental in classical image processing for the detection of edges and corners, where spatial derivatives of intensity functions are utilized to identify discontinuities that correspond to structural features such as object boundaries and junctions. Operators such as Sobel and Prewitt compute finite differences through convolutional kernels that approximate local gradients along horizontal and vertical directions \cite{gonzalez2008digital}. These kernels not only emphasize high-frequency components but also maintain spatial coherence, thus enhancing performance in subsequent tasks such as image segmentation and feature correspondence \cite{canny}\\
\noindent To perform quantum-based edge and corner detection, we first require a mechanism to estimate the gradient of an image encoded in a quantum state. This section introduces a novel quantum circuit—referred to as the \emph{Quantum Gradient Kernel Circuit}—which efficiently computes a \emph{lag-2 difference}, which forms the core primitive for Sobel-style gradient estimation.

\noindent The lag-$k$ difference of an image vector \( I_{\text{vec}} \) at position \( i \) is defined as Equation \ref{eq8}:
\begin{equation}
\Delta_k(i) = I_{\text{vec}}(i + k) - I_{\text{vec}}(i - k)   
\label{eq8}
\end{equation}

\noindent where \( k \in \mathbb{N} \) controls the stride of differencing. In particular, the \emph{lag-2 difference} (\( k=2 \)) captures more spatial context than the lag-1 difference and is critical in approximating the behavior of classical Sobel operators, which emphasize central pixel gradients using a weighted scheme.

\noindent We design a quantum circuit that implements this lag-2 differencing using a combination of quantum memory access and controlled operations. Compared to existing quantum circuits that implement lag-1 difference operations such as the Quantum Harris Edge Detector (QHED), our circuit requires only a polynomial overhead in the number of qubits \( r \), i.e., it uses \( \text{poly}(r) \) additional gates. This makes the proposed circuit both scalable and practical for NISQ-era implementations.

\noindent The resulting gradient information can then be passed to downstream quantum or classical modules for edge detection (via thresholding) and corner detection (e.g., via Quantum Harris Corner Detection).

\subsubsection*{Computing lag-2 difference using QPIE}\label{subsubsec3} 
Consider a 2D image $I$ encoded as following Equation \ref{eq9}
\begin{equation}
\vert I \rangle = \sum_{i=0}^{N-1} c_i \vert i \rangle \equiv [c_0,c_1,\cdots,c_{N-1}]^{\texttt{T}}   
\label{eq9}
\end{equation}

\noindent where $r$ denotes the number of position qubits and $N = 2^r$ is the number of encoded pixel positions, and $c_i$ is the normalized pixel intensity value of $i$-th pixel. We then introduce an ancilla qubit to $\vert I \rangle$ and apply the Hadamard gate on the ancilla qubit. This results in a $(r+1)$-qubit quantum state of the form like Equation \ref{eq10}: 
\begin{equation}
\vert \Bar{I} \rangle := \vert I \rangle \otimes \vert + \rangle = \tfrac{1}{\sqrt{2}}[c_0,\ c_0,\ c_1,\ c_1, \cdots, c_{N-1}, \ c_{N-1}]^{\texttt{T}}
\label{eq10}
\end{equation}
\noindent Next, we perform a unitary operator for $(r+1)$-qubit amplitude permutation, denoted as $\mathcal{D}_{2^{(r+1)}}$, which shifts the elements of $\vert \Bar{I} \rangle$ down by one position, producing the state as Equation \ref{eq11}:

\begin{equation}
\mathcal{D}_{2^{(r+1)}}\vert \Bar{I} \rangle = \tfrac{1}{\sqrt{2}}[c_0,\ c_1,\ c_1,\ c_2, \cdots, c_{N-1}, \ c_{N-1}, \ c_0]^{\texttt{T}}
\label{eq11}
\end{equation}

where $$\mathcal{D}_{2^{(r+1)}}:= \begin{bmatrix}
    0 & 1 & 0  & \cdots & 0\\
    0 & 0 & 1  & \cdots & 0\\
    \vdots & \vdots & \vdots & \ddots & \vdots\\
    0 & 0 & 0  & \cdots & 1\\
    1 & 0 & 0  & \cdots & 0
\end{bmatrix}$$

\noindent This shift operator can be efficiently implemented in $O(r)$ quantum gates, where $r$ denotes the number of position qubits \cite{fijany1999quantum}. The operator acts on an $(r+1)$-qubit system, including the ancilla qubit. The cyclic shift corresponds to an addition modulo $2^r$ on the position register, which can be efficiently realized using standard quantum adder circuits (e.g., ripple-carry adders) with $O(r)$ elementary gates. This formulation ensures that the complexity is expressed in terms of the number of qubits rather than the number of encoded pixel positions $(N = 2^r)$, thereby avoiding any ambiguity between qubit-based and pixel-based complexity measures.

\begin{figure}[!htb]
    \centering
    \includegraphics[width=0.95\linewidth]{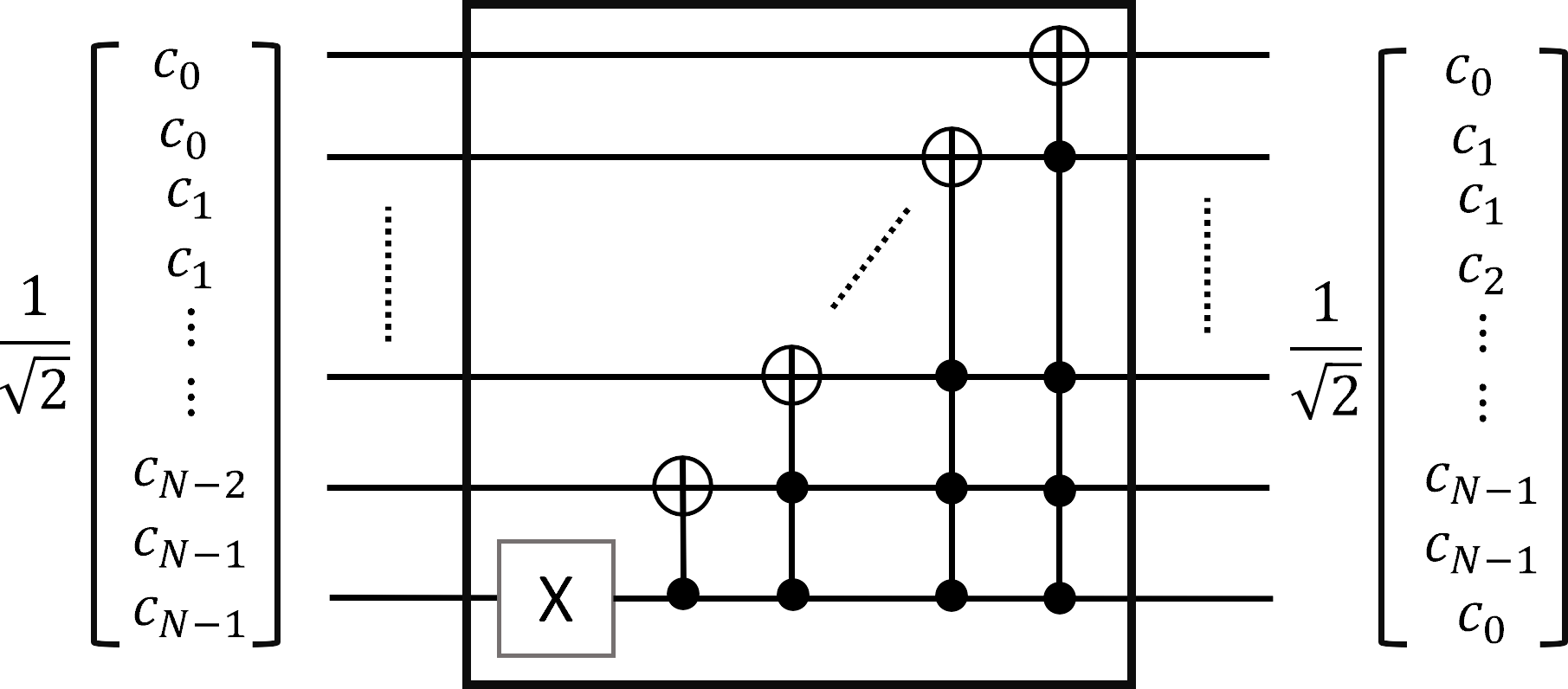}
    \caption{Quantum circuit for implementing $(r+1)$-qubit amplitude permutation unitary operator $\mathcal{D}_{2^{(r+1)}}$.}
    \label{fig:shiftoperator}
\end{figure}

A quantum circuit for implementing the $(r+1)$-qubit shift operator $\mathcal{D}$ (of dimension $2^{r+1} \times 2^{r+1}$) using multi-control Pauli-$X$ gates is illustrated in Figure~\ref{fig:shiftoperator}.

\noindent We then apply a Pauli-$X$ gate on the ancilla qubit, leading to the swap of elements in non-overlapping consecutive blocks of two elements. This results in the state like Equation \ref{eq12}

\begin{equation}
\begin{aligned}
(I^{\otimes r}\otimes X)\mathcal{D}_{2^{(r+1)}} \vert \Bar{I} \rangle = \tfrac{1}{\sqrt{2}}[c_1,\ c_0,\ c_2,\ c_1, \ c_3, \ c_2, \\
\quad \cdots,\ c_{N-2}, \ c_{0}, \ c_{N-1}]^{\texttt{T}}   
\label{eq12}
\end{aligned}
\end{equation}

\noindent We again apply the shift operator $\mathcal{D}_{2^{(r+1)}}$ to get the following state as Equation \ref{eq13}

\begin{equation}
\begin{aligned}
    \mathcal{D}_{2^{(r+1)}}(I^{\otimes r}\otimes X) \mathcal{D}_{2^{(r+1)}} \vert \Bar{I} \rangle = \tfrac{1}{\sqrt{2}}[c_0,\ c_2,\ c_1,\ c_3,\ c_2, \\
\quad \cdots,\ c_{N-2},\ c_{0},\ c_{N-1},\ c_1]^{\texttt{T}} 
\label{eq13}
\end{aligned}
\end{equation}

\noindent Finally, to obtain the lag-2 difference for all pixels, i.e., $(c_0-c_2),(c_1-c_3),(c_2-c_4), \cdots$, we apply the Hadamard gate on the ancilla qubit, resulting in a state of the form Equation \ref{eq14}:

\begin{equation}
\begin{aligned}
(I^{\otimes r} \otimes H)\mathcal{D}_{2^{(r+1)}}(I^{\otimes r} \otimes X)\mathcal{D}_{2^{(r+1)}} \vert \Bar{I} \rangle 
= \tfrac{1}{2} \big[ (c_0 + c_2), \ (c_0 - c_2), \\
\ (c_1 + c_3), \ (c_1 - c_3), \ (c_2 + c_4),\ (c_2 - c_4),\  \cdots, \\ \ (c_{N-2} + c_0), \ (c_{N-2} - c_0), (c_{N-1} + c_1),\ (c_{N-1} - c_1) \big]^{\texttt{T}}
\label{eq14}
\end{aligned}
\end{equation}

\noindent\textbf{Recovering true pixel-intensity differences from lag-2 amplitude differences.}
Equation (19) outputs lag-2 differences in the \emph{normalized amplitude domain}.
Let $\Delta^{(c)}_{2}(k)$ denote the (unnormalized) amplitude difference
\begin{equation}
\Delta^{(c)}_{2}(k) := c_k - c_{k+2},
\end{equation}
(where $k+2$ follows the same indexing convention used in the circuit).
Because $c_k=I_{vec}(k)/\|I_{vec}\|_2$, the corresponding lag-2 difference in the
\emph{original pixel-intensity domain} is obtained by the direct rescaling:
\begin{equation}
\Delta^{(I)}_{2}(k) := I_{vec}(k)-I_{vec}(k+2) = \|I_{vec}\|_2 \,\Delta^{(c)}_{2}(k).
\end{equation}
Hence, Sobel-derived horizontal/vertical gradients computed from lag-2 differences
must be scaled by $\|I_{vec}\|_2$ to match the absolute-gradient convention used in
classical image processing.
This scaling also propagates to the Harris structure tensor $M$ and score $R(M)$:
if $I_x$ and $I_y$ are multiplied by $\|I_{vec}\|_2$, then $M$ scales by $\|I_{vec}\|_2^2$
and the response scales by $\|I_{vec}\|_2^4$ (since $R(M)=\det(M)-\kappa\mathrm{Tr}(M)^2$).
Accordingly, in our numerical benchmarks, we either rescale gradients back to the
pixel domain or equivalently normalize the Harris response threshold consistently
across methods so that absolute corner scores remain comparable.\\
Importantly, multiplying all gradient components by a single global scaling factor does not change the locations of local maxima of the Harris response function $R(M)$; it only affects the absolute magnitude of the response values. Therefore, the detected corner positions remain invariant under such scaling, provided that a consistent thresholding strategy is applied.T his property ensures that the comparison between quantum and classical corner detection remains valid in terms of spatial localization, even though their response magnitudes may differ due to normalization.

\noindent Thus, by measuring the ancilla qubit, conditioning on obtaining $1$, we obtain the $r$-qubit state as Equation \ref{eq15}
\begin{equation}
 \tfrac{1}{2}[(c_0-c_2), \ (c_1-c_3),\ (c_2-c_3), \cdots, (c_{N-2}-c_2),\  (c_{N-1}-c_1)]^{\texttt{T}}  
\label{eq15}
\end{equation}

\noindent which contains the lag-2 difference for all pixels along $x$ direction. For lag-2 difference in the $y$ direction, the image is first transposed, and the same steps are repeated.

\subsubsection*{Computing lag-2 difference using FRQI}\label{subsubsec4}
Consider a 2D image $I$ encoded using Equation \ref{eq16} 

\begin{equation}
 \ket{I}=\frac{1}{\sqrt{2^r}} \sum_{i=0}^{2^{r}-1}(\cos{{\theta}_i}\ket{0}+\sin{{\theta}_i}\ket{1}) \otimes \ket{i}
\label{eq16}
\end{equation}

\noindent Observe that the pixel intensity value is encoded in the ancilla qubit. If we perform a measurement on the ancilla qubit in the Pauli-$Z$ basis, we get a quantum image representation similar to QPIE, as described below \\

\noindent If outcome = 0:
\begin{equation}
    \ |I_c\rangle := \sum_{i=0}^{2^r-1} \frac{\cos(\theta_i)}{\sqrt{\sum_j \cos^2(\theta_j)}} |i\rangle.
    \label{eq17}
\end{equation} 
If outcome = 1:
\begin{equation}
    \ |I_c\rangle := \sum_{i=0}^{2^r-1} \frac{\sin(\theta_i)}{\sqrt{\sum_j \sin^2(\theta_j)}} |i\rangle.
\label{eq18}
\end{equation}

\noindent Keeping this in mind, we can apply the same technique used for computing lag-2 difference using QPIE. However, in this case, we compute the differences in terms of sine and cosine of pixel intensity values.

\begin{figure}[!htb]
    \centering
    \includegraphics[scale=0.25]{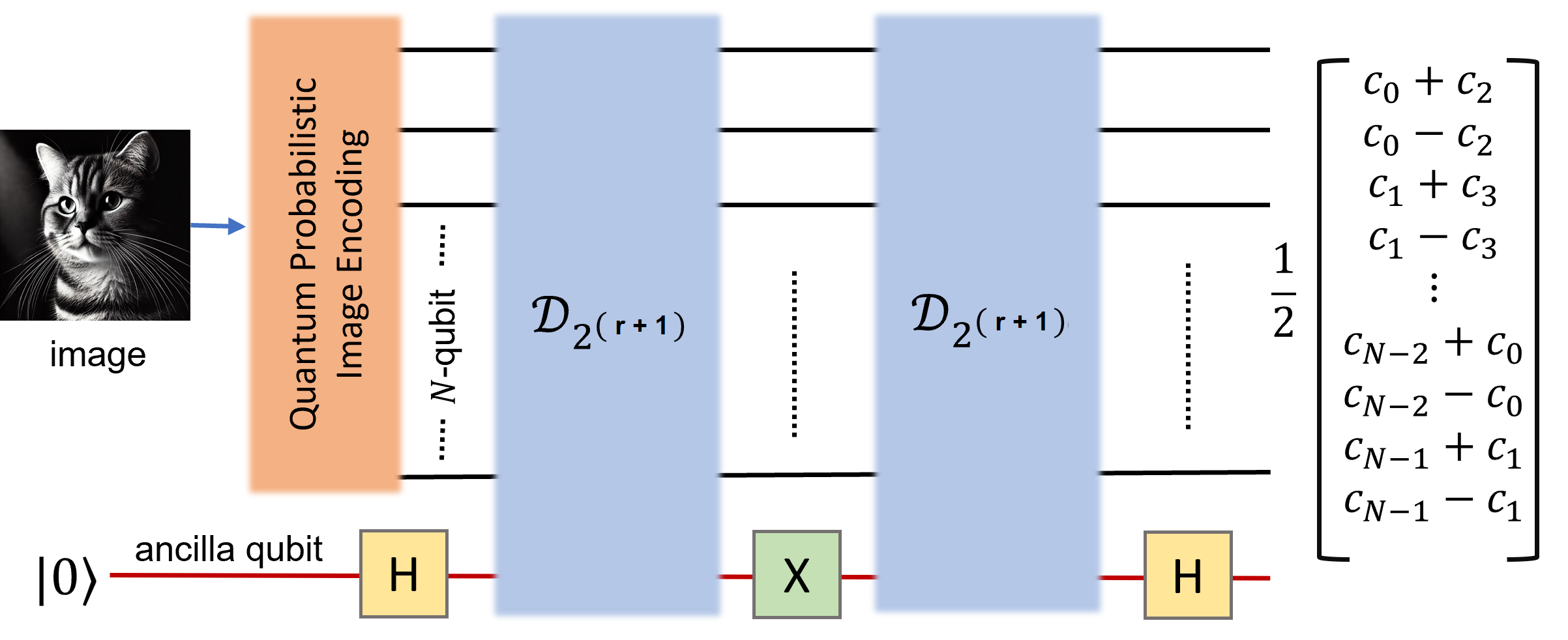}
    \caption{Quantum circuit for computing lag-2 difference using QPIE method.}
    \label{fig:QPIE_lag2}
\end{figure}

\begin{figure}[!ht]
    \centering
    \includegraphics[scale=0.2]{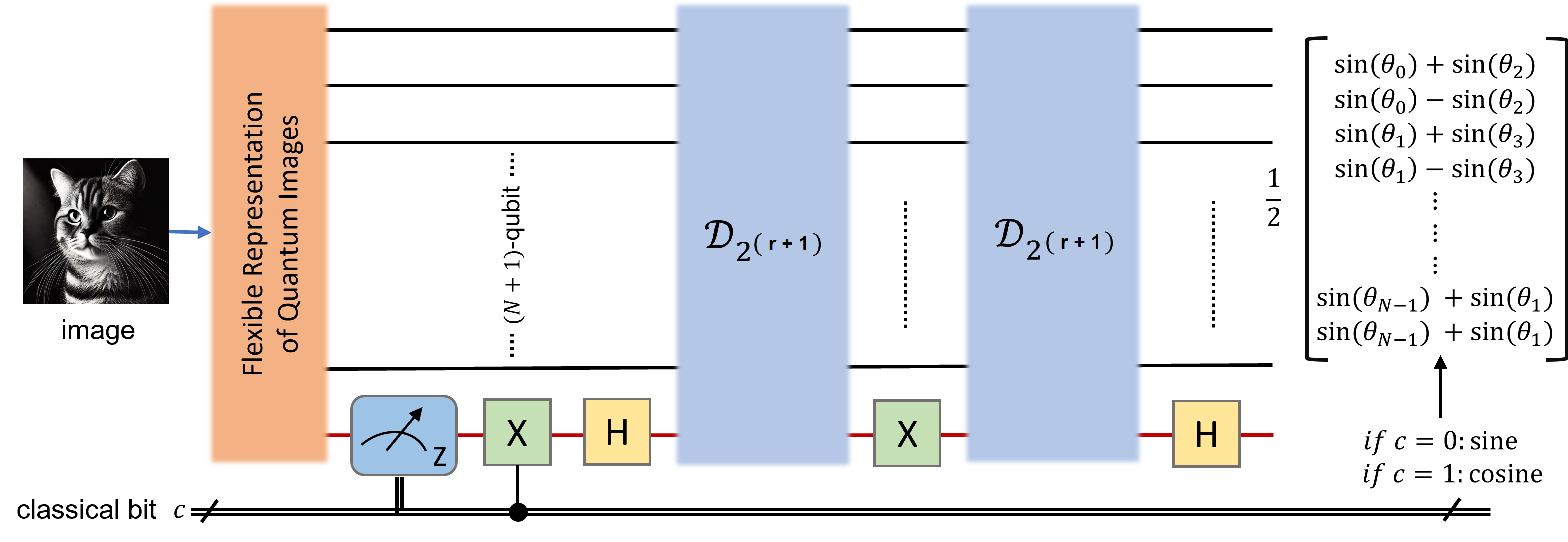}
    \caption{Quantum circuit for finding lag-2 difference using FRQI method.}
    \label{fig:FRQI_lag2}
\end{figure}
The quantum circuits for the lag-2 difference calculation using QPIE and FRQI are illustrated in Figure ~\ref{fig:QPIE_lag2} and ~\ref{fig:FRQI_lag2}, respectively. The proposed lag-2 difference operation is implemented via a sequence of controlled amplitude permutations acting on the position register. Each permutation corresponds to a cyclic shift or swap operation that can be decomposed into a polynomial number of elementary quantum gates. Specifically, for an image encoded using \( r \) position qubits, the required permutation operations can be implemented with \( \mathcal{O}(r) \) controlled-SWAP and controlled-NOT gates, resulting in an overall gate complexity that scales polynomially with the input size. Therefore, the lag-2 difference computation does not introduce exponential overhead and remains efficient within the proposed quantum pipeline. 
\subsection{Classical Post-Processing for Sobel Kernel}\label{subsec4}
After obtaining the lag-2 difference along both the $x$ and $y$ directions, we proceed with classical post-processing to compute the horizontal and vertical gradients, as calculated using the Sobel kernel. To illustrate this, consider a 2D image $I = [I_{i,j}]$. The horizontal gradient corresponding to the Sobel kernel $G_x$ for a pixel located at $(i,j)$ is calculated using Equation \ref{eq19}:

\begin{equation}
I_x(i,j) = (I_{i-1,j-1} - I_{i+1,j-1})   + 2\cdot(I_{i-1,j} - I_{i+1,j}) +
(I_{i-1,j+1} - I_{i+1,j+1})   
\label{eq19}
\end{equation}

The vertical gradient at a specific pixel, using the Sobel kernel \( G_y \), is calculated as the following Equation \ref{eq20}:

\begin{equation}
I_y(i,j) = (I_{i-1,j-1} - I_{i-1,j+1}) + 2 \cdot (I_{i,j-1} - I_{i,j+1}) + (I_{i+1,j-1} - I_{i+1,j+1})
\label{eq20}
\end{equation}

\noindent The differences used in these formulas can be accurately derived from the lag-2 differences computed using the quantum circuit up to a normalization constant.

\subsection{Edge Detection}\label{subsec5}
In this section, we present our QHED algorithm. We begin with a brief overview of classical Sobel edge detection to establish a baseline for evaluating the quantum-enhanced method.
\subsubsection{Sobel Edge Detection}\label{subsubsec5}
The Sobel operator yields more continuous and coherent edges, effectively highlighting structural boundaries within the image. To detect edges in an image, one typically computes the intensity gradient at each location of the pixels. Given a grayscale image represented as a two-dimensional function $I(x, y)$, the partial derivatives with respect to the directions $x$ and $y$  are calculated by Equations \ref{eq21}.
\begin{equation}
I_x = \frac{\partial I}{\partial x},  \ \ \ \ \  I_y = \frac{\partial I}{\partial y}  
\label{eq21}
\end{equation}
\noindent In practice, these derivatives are approximated using discrete convolution operators such as the Sobel kernel. Once $I_x$ and $I_y$ are computed, the magnitude of the gradient in each pixel is given in Equation \ref{eq22}:

\begin{equation}
\|\nabla I(x, y)\| = \sqrt{I_x(x, y)^2 + I_y(x, y)^2}
\label{eq22}
\end{equation}

\noindent This magnitude measures the strength of the edge at each point. Optionally, the gradient direction can also be computed using Equation \ref{eq23}:

\begin{equation}
\theta(x, y) = \arctan\left(\frac{I_y(x, y)}{I_x(x, y)}\right)
\label{eq23}
\end{equation}

\noindent Pixels with large gradient magnitude values are considered edge candidates. To obtain a binary edge map, a thresholding operation is applied as defined in Equation \ref{eq24}:

\begin{equation}
E(x, y) = 
\begin{cases}
1 & \text{if } \|\nabla I(x, y)\| \geq \tau \\
0 & \text{otherwise}
\end{cases}
\label{eq24}
\end{equation}

\noindent where $\tau$ is a predefined threshold. This process effectively highlights regions in the image where the intensity changes sharply, indicating the presence of edges.

\subsubsection{Quantum Hadamard Edge Detection (QHED)}\label{subsubsec6}
In this study, we propose a QHED algorithm that leverages quantum parallelism to enhance the process of identifying intensity discontinuities in digital images by building on the classical framework of gradient-based edge detection. QHED algorithm utilizes the Hadamard gate to distribute information across superposed quantum states, thereby encoding high-frequency components associated with edges. The core idea is to perform a lag-2 difference computation within a quantum circuit and apply Hadamard transformations on the amplitude-encoded image data. This process causes constructive and destructive interference that amplifies the presence of abrupt intensity changes while suppressing uniform or slowly varying regions. The QHED tends to produce fragmented or disconnected edge representations, often resulting in numerous edge segments rather than smooth contours. Sharp spatial transitions manifest as high probability amplitudes, which can be classically thresholded to generate binary edge maps. Compared to classical operators such as Sobel, the QHED circuit is compact and hardware-efficient, as it avoids costly convolution operations and instead exploits quantum interference to approximate derivative behavior. It should be noted that the original QHED algorithm proposed by \cite{ref7} is based on a lag-1 differencing scheme. In contrast, the approach adopted in this work employs a lag-2 differencing strategy within the Quantum Gradient Kernel Circuit described in Section 2.3. Therefore, the method used here can be considered as a modified version of QHED, and all experimental results in this study are based on this lag-2 formulation.

\subsection{Corner Detection}\label{subsec6}

In this section, we present our QHCD algorithm. We begin with a brief overview of the (classical) Harris corner detection algorithm. 

\subsubsection{Harris corner detection (HCD)}\label{subsubsec7}
HCD algorithm, developed by Chris Harris and Mike Stephens in 1988, is a widely used method to detect corners and features in images \cite{ref9}. Unlike earlier methods that relied on shifting patches across different fixed angles \cite{moravec1980}, HCD uses a differential approach to compute a \textit{corner response function} directly by considering intensity variations across multiple directions. The Harris algorithm is robust against changes in image scale, rotation, and noise, which makes it suitable for a variety of computer vision applications \cite{schmid2000evaluation}.

\noindent Consider a pixel block $B_I$ inside a 2D image $I$. The HCD algorithm identifies corners in $B_I$ by maximizing the following corner response function (Equation \ref{eq26}). 

\begin{equation}
     f(\Delta_x,\Delta_y)\! := \!\!\!\!\!\!\sum_{(x,y)\in B_{I}} \!\!\!w(x,y) [I(x+\Delta_x,y+\Delta_y)-I(x,y)]^2
     \label{eq26}   
\end{equation}

\noindent where \(w(x,y)\) represents the window function (e.g., a rectangular or Gaussian window) that assigns weights to pixels of $B_I$, \(I(x, y)\) represents image intensity at pixel $(x,y)$, and  \(I(x+\Delta_x, y+\Delta_y)\) represents the intensity of the image at a shifted pixel. Let \( I_x \) and \( I_y \) be the partial derivatives of $I$, such that the intensity at a nearby point \((x + \Delta x, y + \Delta y)\) can be approximated by Equation \ref{eq27}:
\begin{equation}
 I(x + \Delta_x, y + \Delta_y) \approx I(x, y) + I_x(x, y)\, \Delta_x + I_y(x, y)\, \Delta_y
\label{eq27}    
\end{equation}
\noindent By applying the Taylor series expansion to the above equation, we obtain Equation \ref{eq28}:

\begin{equation}
    f(\Delta_x,\Delta_y)\approx \left[\Delta_x\ \Delta_y\right]M\left[\begin{matrix}\Delta_x\\\Delta_y\\\end{matrix}\right]
    \label{eq28}
\end{equation}

\noindent where $$ M := \sum_{x,y\in B_{I}} w(x,y) \left[\begin{matrix} I_x^2 & I_x I_y\\ I_x I_y& I_y^2\\\end{matrix}\right],$$ and  $I_x$ and $I_y$ are image gradients along $x$ and $y$ directions, respectively. The matrix $M$ essentially captures the intensity variation along $x$ and $y$ directions inside $B_I$. The image gradients can be computed using different convolution masks, such as the Sobel kernel \cite{vincent2009descriptive}. Specifically, the partial derivatives of the image intensity, \( I_x \) and \( I_y \), are obtained by convolving the image \( I \) with the Sobel kernels \( G_x \) and \( G_y \) with Equations \ref{eq29}, respectively:
\begin{equation}
I_x = G_x * I \ \text{ and } \  I_y = G_y * I
\label{eq29}
\end{equation}
where \( * \) denotes the convolution operation and \( G_x \) and \( G_y \) are $3\times 3$ Sobel kernels defined as Equations \ref{eqm}:
\begin{equation}
G_x = \begin{bmatrix} \ -1 & \ 0 & \ 1 \\ \ -2 & \ 0 & \ 2 \\ \ -1 & \ 0 & \ 1 \end{bmatrix} \ \text{ and } \
G_y = \begin{bmatrix} \  -1 & \  -2 & \ -1 \\ \ \    0 & \ \   0 & \ \   0 \\   \ \ 1 & \ \  2 &  \ \ 1 \end{bmatrix}.
\label{eqm}
\end{equation}
By evaluating these gradients and subsequently constructing the matrix \( M \), the algorithm distinguishes between flat regions, edges, and corners. To formally determine whether a corner exists within the block \( B_I \), the HCD algorithm uses a score function $\mathcal{R}(M)$ defined by Equation \ref{eq30}:
\begin{equation}
     \mathcal{R}(M) := \det(M) - \kappa \, \Tr(M)^2
    \label{eq30} 
\end{equation} 

\noindent where \(\kappa\) is a hyperparameter that controls the sensitivity of the corner detection. Since \(M\) is a \(2 \times 2\) matrix, its determinant and trace can be calculated using its eigenvalues. Let $\lambda_1$ and $\lambda_2$ be the eigenvalues of $M$. 
Then, the score function \( \mathcal{R} \) can be re-written as Equation \ref{eq31}:
\begin{equation}
\mathcal{R}(\lambda_1,\lambda_2) = (\lambda_1\lambda_2) - \kappa\,(\lambda_1+\lambda_2)^2
    \label{eq31} 
\end{equation} 
Therefore, the magnitudes of these eigenvalues classify the region within \( B_I \) as
\begin{itemize}
\item \textit{Flat region}: If \( \lvert \mathcal{R} \rvert \) is small (less than a determined threshold involving \(\lambda_1\) and \(\lambda_2\)), the block \( B_I \) represents a {flat region} with minimal intensity variation.
\item \textit{Edge}: If \( \mathcal{R} < 0 \), the block \( B_I \) represents an {edge}, indicating intensity variation in only one direction.
\item \textit{Corner}: If \( \lvert \mathcal{R} \rvert \) is large and positive, it indicates that the block \( B_I \) contains a {corner}, as there are significant intensity variations in multiple directions.
\end{itemize}
Finally, the result of HCD is a grayscale image with scores calculated for each pixel block $B_I$. After applying a suitable threshold to the scores, we can identify the corners in the image $I$. 
In HCD, the algorithm relies heavily on gradient computation using a convolution mask. For an image of size \(2^n\times2^n\), the classical convolution requires \(O(2^{2n})\) operations, as each of the \(2^{2n}\) pixels must be processed individually.

\noindent To address this computational challenge, we propose using a quantum circuit to perform the gradient calculation, which provides a theoretical speedup at the circuit level for the gradient-computation subroutine. In the next subsection, we discuss our QHCD algorithm and present the quantum circuit designed to compute gradients efficiently, focusing on the Sobel kernels.

\subsubsection{Quantum Harris corner detection (QHCD)}\label{subsubsec8}
In QHCD, we first convert a classical image into its quantum representation using techniques such as QPIE and FRQI. Next, we compute the image gradients along both the horizontal $x$ and vertical $y$ directions using the Quantum Gradient Kernel Circuit introduced in Section 2.3. This quantum-based gradient computation is a key differentiator: the quantum circuit computes lag-2 differences for all pixel positions in superposition using only \(O(\text{poly}(n))\) quantum gates. Finally, similar to traditional HCD, we perform classical post-processing on the computed gradients to accurately detect the corners in the image.

\noindent The quantum advantage claimed in this work should be interpreted with care and refers specifically to \emph{circuit-level parallelism} in the gradient computation stage: enabled by quantum superposition, all pixel positions are processed simultaneously within a single quantum state during one coherent circuit execution, using a number of quantum gates that scales polynomially with the number of qubits. However, as in other amplitude-encoded schemes such as QPIE, recovering the full gradient map in classical form requires repeated circuit evaluations, and both state preparation and measurement scale as \(O(2^{2n})\) in the worst case. Consequently, the quantum parallel advantage is realized at the circuit level through simultaneous lag-2 difference evaluation, while the subsequent extraction of gradient values is carried out through classical post-processing consistent with amplitude-encoded quantum image representations, and the overall hybrid quantum--classical pipeline does not achieve an end-to-end exponential speedup over classical methods. Nevertheless, the proposed framework highlights a meaningful separation between theoretically efficient quantum subroutines and current practical limitations imposed by NISQ-era hardware, providing a foundation for future advances in quantum image processing as state preparation and readout techniques improve.

\noindent The central idea for gradient computation using the Sobel kernels is based on the lag-2 difference between pixel intensities, i.e., $(I_{i,j} - I_{i,j+2})$ along $x$ direction and $(I_{i,j} - I_{i+2,j})$ along $y$ direction, where $I_{i,j}$ is the pixel intensity value of the $(i,j)$-th pixel. After post-processing, we can calculate the gradient using these differences. For instance, consider a $3\times 3$ (normalized) image, 
\begin{align*}
    I = & \begin{bmatrix}
    \ \ c_0 & \ \ c_1 & \ \ c_2\\
    \ \ c_3 & \ \ c_4 & \ \ c_5\\
    \ \ c_6 & \ \ c_7 & \ \ c_8\\
 \end{bmatrix} \overset{\text{QPIE}}{\longrightarrow} \ |I\rangle\! =\! [c_0, c_1, c_2,\cdots,c_6, c_7, c_8]^{\texttt{T}}.
\end{align*}

The gradient in pixel $(2,2)$ along $x$ direction using the Sobel kernel $G_x$ can be written as Equation \ref{eq32} 
\begin{equation}
    I_x(2,2) =(c_0-c_2) + 2(c_3-c_5) + (c_6-c_8)
    \label{eq32}
\end{equation}
Thus, by calculating the differences $(c_{0}-c_2), (c_3-c_5),$ and $(c_6-c_8)$, we can appropriately scale these values and combine them to obtain the gradients along the \(x\) direction. Similarly, to obtain the gradient along the \(y\) direction, we can exploit the relationship between the Sobel kernels: \(G_y = G_x^{\texttt{T}}\). This means that by transposing the image \(I\) and applying \(G_x\), we effectively calculate the vertical differences $(c_0 - c_6), (c_1 - c_7), \text{ and } (c_2 - c_8)$. In light of this, we conclude that if we calculate the lag-2 difference using a quantum circuit, then we can effectively compute the gradient along both \(x\) and \(y\) directions corresponding to the Sobel kernels. To begin with, we first outline the design of the quantum circuit for computing the lag-2 difference based on QPIE as its underlying encoding mechanism. We further extend the circuit for the FRQI encoding method based on the architecture proposed in \cite{ref13}. \\

\noindent \textbf{Using SOBEL Kernel}
Following this, we adopt a procedure similar to the HCD. First, we compute the matrix $M$ for a pixel block $B_I$ within the image $I$. Then, we evaluate the score function $\mathcal{R}(M)$ to identify the corner within the block $B_I$. It is important to note that, in the case of QPIE, we get the normalized pixel intensity values, whereas in FRQI, we get the sine or cosine of the pixel intensity values. Therefore, we must carefully select the hyper-parameters and thresholds needed to determine the corners effectively.

\begin{figure}[!htb]
  \centering
  \begin{subfigure}[t]{0.43\linewidth}
\centering
\setlength{\fboxsep}{0pt}
\fbox{\includegraphics[width=0.58\linewidth]{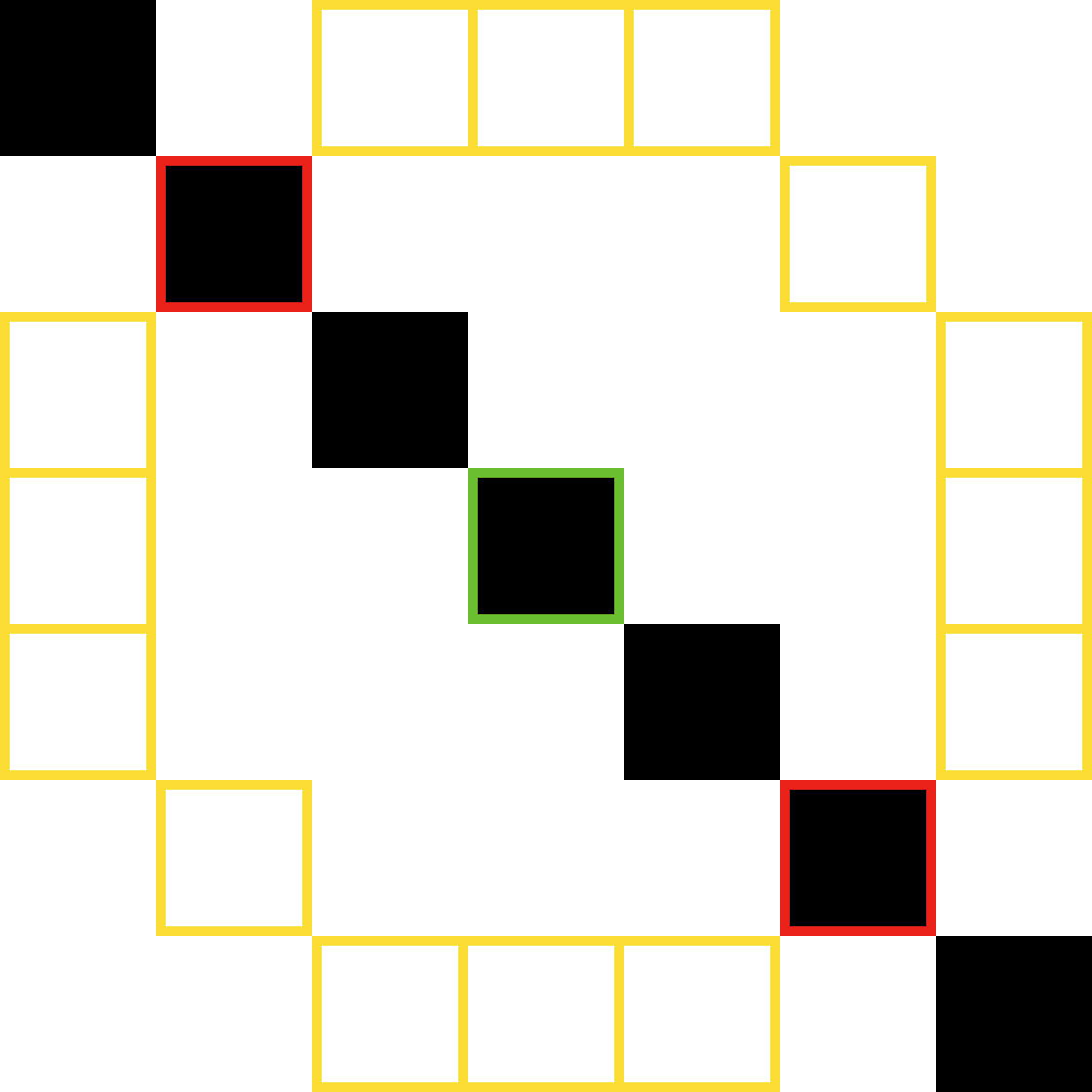}}
    \caption{Wrongly detected corner}
    \label{edge}
  \end{subfigure}\quad\quad
  \begin{subfigure}[t]{0.44\linewidth}
\centering
\setlength{\fboxsep}{0pt}
\fbox{\includegraphics[width=0.56\linewidth]{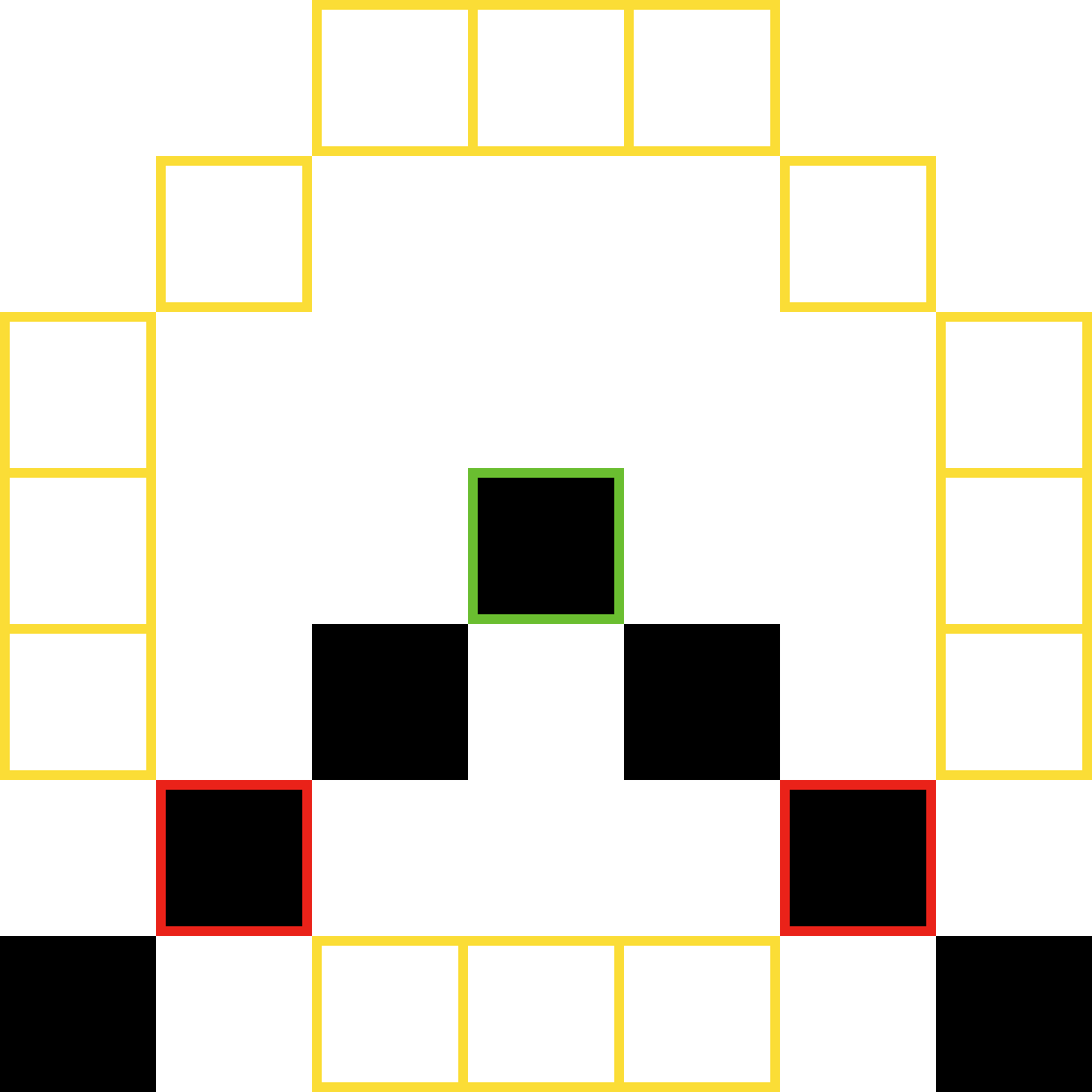}}
    \caption{Correctly detected corner}
    \label{corner}
  \end{subfigure}
    \caption{
Example of a wrongly and a correctly detected corner. In both images, the suspected corner is highlighted in green, the border of a 3-pixel circle around it is highlighted in yellow, and the pixels along the border are highlighted in red. 
In the first image, there are two black pixels located on opposite sides of the border, indicating that the suspected corner was incorrectly detected; it is actually an edge pixel. In the second image, there are also two pixels on the border, but they are not positioned on opposite sides, which indicates that the corner was correctly detected.}
   \label{fig5} 
\end{figure}

\noindent To enhance the quality of the algorithm, an additional post-processing step is applied to each point identified as a corner by the QHCD algorithm. Drawing on the FAST corner detection method \cite{fastcorner}, a point in the binary output of the full derivative is considered to correspond either to a corner or to an edge. If any pixel on the perimeter of a circular neighborhood with radius \( r = 2 \) surrounding the suspected corner has a value of ``1'', and at least one of the two most distant pixels on this circle also has a value of ``1'', then the suspected corner is interpreted as an edge that has been incorrectly classified as a corner and is therefore discarded. \\
\noindent Here, the term \emph{opposite pixels} refers to pairs of pixels located diametrically opposite to each other on the verification circle, i.e., separated by 180° along the same radial direction. Figure~\ref{fig5} illustrates examples of both a wrongly detected corner and a correctly detected corner. The radius \( r = 2 \) is selected as a compromise between spatial locality and robustness to noise, providing sufficient spatial support for corner validation while remaining consistent with the scale of Sobel-based gradient estimation.

\section{Experiments and Results}\label{sec3}

In all FRQI-based experiments presented in this section, we adopt the linear FRQI encoding defined in Equation~\ref{eqn:frqi_enc2}. This choice is motivated by its ability to preserve relative intensity differences, which are essential for gradient-based edge and corner detection using the Sobel kernel. Nonlinear (arcsin-based) FRQI encoding is therefore not considered in the experimental evaluation, as it introduces nonlinear intensity distortion that may obscure gradient magnitude relationships required for accurate Sobel-based detection. In order to test the efficiency of the proposed Quantum Harris Corner algorithm, a series of experiments comparing it to the original QHED algorithm were performed. The experiments were divided between edge and corner detection to individually analyze their performance in each use case. 

\noindent\textbf{Experimental setup.}
All edge and corner detection experiments are conducted on images resized to a resolution of $512 \times 512$ using classical preprocessing implemented with OpenCV (\texttt{cv2}). Under this resolution, the QPIE representation requires $r=\lceil \log_2(512^2) \rceil = 18$ qubits, while the FRQI representation requires $2\log_2(512)+1=19$ qubits. All quantum circuits are simulated in an ideal noiseless setting using IBM's simulation environment with the Qiskit framework~\cite{ref17}. For both the QHED method and the quantum gradient (lag-2) based Sobel approach, simulations are performed in a statevector setting without finite-shot sampling, and the results are obtained directly from the amplitudes of the simulated quantum states. All simulations are executed on an NVIDIA A100 GPU.\\
It should be noted that all edge and corner detection experiments presented in this section are conducted under ideal statevector simulation, where measurement noise and finite-shot effects are not considered. Therefore, the reported results reflect performance in an infinite-shot regime. The impact of finite-shot sampling is analyzed separately in the image reconstruction experiments (Figures \ref{fig:QPIE_FRQI_comparison}, \ref{fig:relDiff_CAT}, \ref{fig:ssim_CAT}) and is not directly evaluated for edge and corner detection tasks in this study. While these observations suggest improved robustness of QPIE under limited measurement shots in the reconstruction setting, this advantage is not explicitly validated for edge and corner detection. A systematic evaluation of finite-shot effects on detection performance remains an important direction for future work.

\subsection*{Edge Detection}\label{subsec7}
\begin{figure}[!htb]
    \centering
    \includegraphics[scale=0.09]{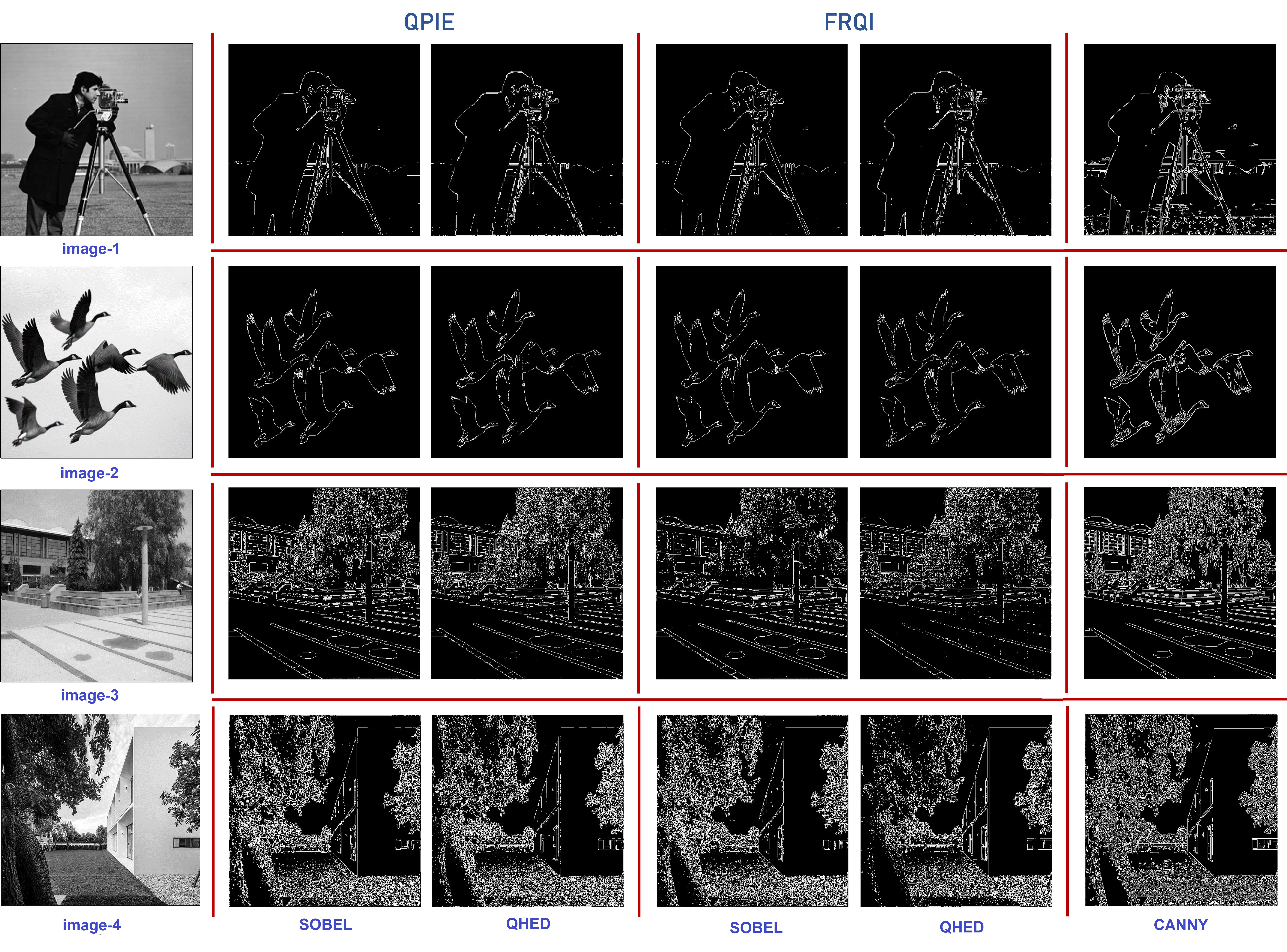}
    \caption{Edge detection results on selected images from the UDED dataset \cite{soria2023teed}. The first column presents the original grayscale images from four distinct scenes: Cameraman (image-1), Flying Geese (image-2), City Walkway (image-3), and Modern House Exterior (image-4). The subsequent columns show edges generated using two quantum image representations, QPIE and FRQI, combined with two edge detection techniques: Sobel and QHED. For both QPIE and FRQI, the Sobel-based results demonstrate smoother and cleaner edge structures, while QHED tends to introduce more noise. Classical Canny edge detection results are also included for comparison. Canny produces smoother and more continuous edge structures, while QHED tends to introduce more fragmented and noise-sensitive responses.}
    \label{fig:edge_detection}
\end{figure}

\begin{figure}[!htb]
    \centering
    \includegraphics[scale=0.09]{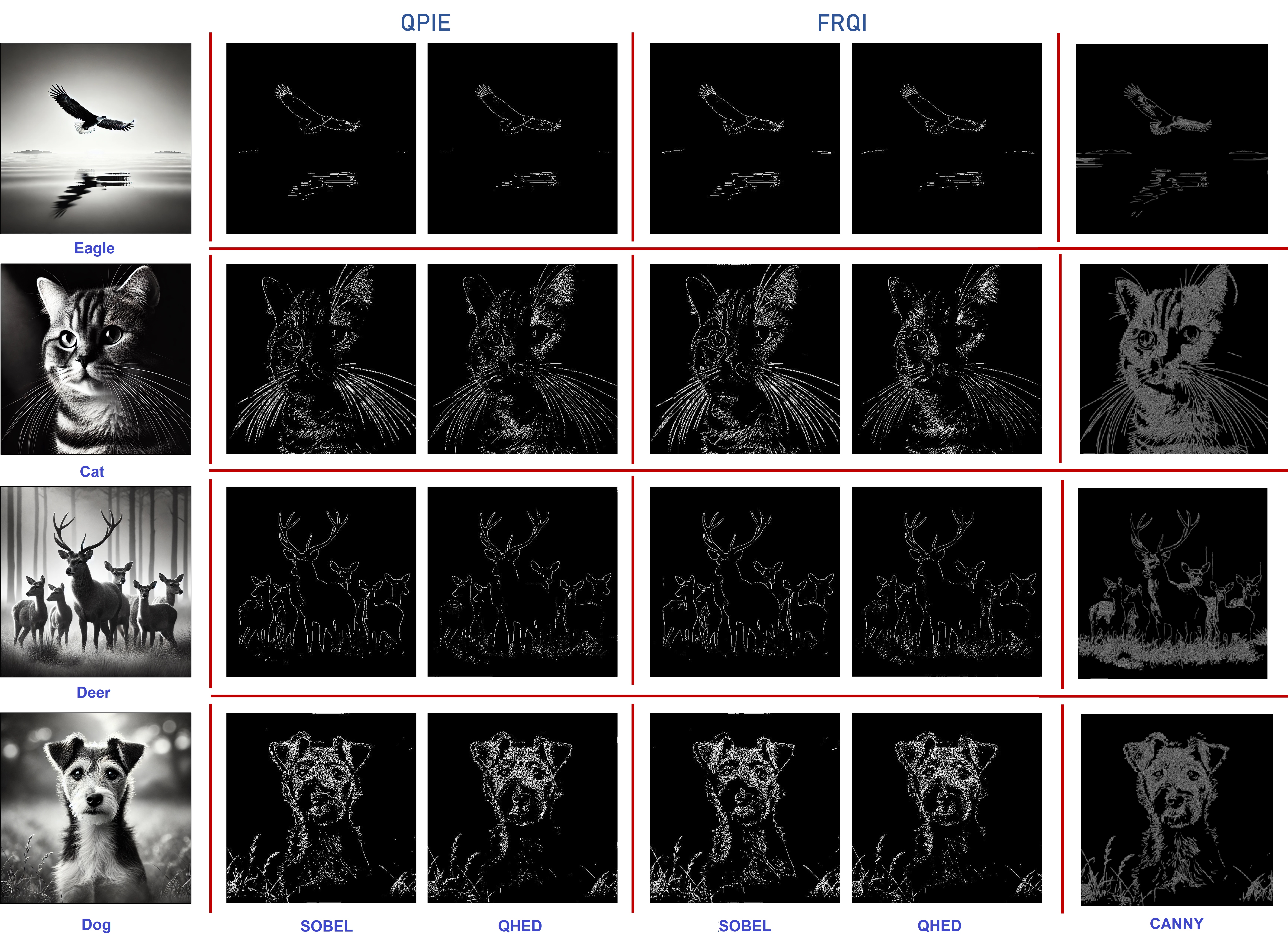}
    \caption{Edge detection results on animal images generated using generative AI. For both QPIE and FRQI, the Sobel-based results demonstrate smoother and cleaner edge structures, while QHED tends to introduce more noise. Classical Canny edge detection results are also included for comparison. Canny produces smoother and more continuous edge structures, while QHED tends to introduce more fragmented and noise-sensitive responses.}
    \label{fig:edge_detection2}
\end{figure}

Four images from the UDED dataset \cite{soria2023teed} were selected to test the edge detection performance of each method, as presented in Figure ~\ref{fig:edge_detection} and ~\ref{fig:edge_detection2}. To quantitatively assess the edge detection quality of the images obtained, we employ four key metrics.\\
\noindent \textbf{Edge Density (ED)} measures the proportion of edge pixels to the total number of image pixels and is computed by Equation \ref{eqED}.

\begin{equation}
\mathrm{ED} = \frac{N_{\text{edge}}}{N_{\text{total}}}
\label{eqED}
\end{equation}
where $N_{\text{edge}}$ is the number of edge pixels and $N_{\text{total}}$ is the total number of pixels in the image. \\

\noindent \textbf{Number of Edge Fragments (EF)} reduction quantifies the relative decrease in EF when using the proposed QHED method compared to the classical Sobel-based approach. It is used to evaluate the effectiveness of QHED in producing more coherent and less fragmented edge maps, where a higher reduction percentage indicates better edge continuity with Equation (\ref{eqEF}).

\begin{equation}
{EF Reduction} (\%) = \left( \frac{\text{QHED EF} - \text{Sobel EF}}{\text{QHED EF}} \right) \times 100
\label{eqEF}
\end{equation}

\noindent \textbf{Edge Thickness (ET)} refers to the average width of edge structures, capturing how sharp or blurred the detected edges appear. 
\begin{equation}
    \mathrm{ET} = \frac{1}{N_{\text{edges}}} \sum_{i=1}^{N_{\text{edges}}} w_i
\label{eqET}
\end{equation}
where \( w_i \) denotes the thickness of the $i--th$ edge structure. \\ 

\noindent \textbf{Edge Entropy (EE)} quantifies the randomness or uncertainty on the edge map, calculated using Equation (\ref{eqEE}).
\begin{equation}
    \mathrm{EE} = -\sum_{i} p_i \log_2 p_i
\label{eqEE}
\end{equation}
where $p_i$ is the probability of the $i$-th intensity level on the edge map.

\begin{table}[h]
\caption{Edge detection performance comparison between Sobel and QHED for QPIE and FRQI image encoding methods. The edge-quality metrics are Edge Density (ED), Edge Thickness (ET), Edge Fragments (EF), and Edge Entropy (EE). `Sobel' and `QHED' refer to classical Sobel-based post-processing applied after quantum encoding and lag-2 difference computation, the modified quantum Hadamard-based edge detection pipeline using the lag-2 formulation, respectively. Overall, under both QPIE and FRQI encodings, classical Sobel-based processing applied to quantum-derived gradients produces more coherent and less noisy edge representations compared to the QHED approach. This observation indicates that the proposed quantum gradient kernel is better suited as a Sobel-like primitive within a hybrid pipeline rather than as a standalone edge detector.}\label{tab:edge}
\centering
\begin{tabular*}{0.90\linewidth}{@{\extracolsep\fill}lccccc}
\toprule%
& & \multicolumn{2}{@{}c@{}}{QPIE} & \multicolumn{2}{@{}c@{}}{FRQI} \\\cmidrule{3-4}\cmidrule{5-6}%
Image & Metric & Sobel & QHED & Sobel & QHED  \\
\midrule
\midrule
\multirow{4}{*}{\rotatebox[origin=c]{0}{image-1}} 
  & ED  & $0.0215$ & $0.0266$ & $0.0216$ & $0.0257$\\
  & ET   & $2.71$   & $1.43$ & $2.77$   & $1.39$ \\
  & \textbf{EF}  & $\textbf{167}$    & $540$ & $\textbf{184}$    & $561$    \\
  & EE & $0.15$   & $0.18$ & $0.15$   & $0.17$   \\
\midrule
\multirow{4}{*}{\rotatebox[origin=c]{0}{image-2}} 
  & ED & $0.0187$ & $0.0198$ & $0.0178$ & $0.0200$ \\
  & ET  & $2.84$   & $1.77$ & $2.76$   & $1.61$  \\
  & \textbf{EF}  & $\textbf{42}$     & $202$ & $\textbf{41}$     & $326$   \\
  & EE & $0.13$   & $0.14$ & $0.13$   & $0.14$   \\
\midrule
\multirow{4}{*}{\rotatebox[origin=c]{0}{image-3}} 
  & ED  & $0.1114$ & $0.1184$ & $0.0788$ & $0.0929$ \\
  & ET & $1.75$   & $1.43$ & $1.86$   & $1.42$   \\
  & \textbf{EF} & $\textbf{1781}$   & $2301$ & $\textbf{1703}$   & $3261$   \\
  & EE  & $0.50$   & $0.52$ & $0.40$   & $0.45$  \\
\midrule
\multirow{4}{*}{\rotatebox[origin=c]{0}{image-4}} 
  & ED  & $0.1959$ & $0.1841$ & $0.2005$ & $0.1396$\\
  & ET  & $1.85$   & $1.68$ & $1.84$   & $1.75$  \\
  & \textbf{EF}  & $\textbf{662}$   & $936$ & $\textbf{794}$   & $1164$   \\
  & EE  & $0.58$   & $0.69$ & $0.72$   & $0.73$  \\
  \midrule
    \midrule
\multirow{4}{*}{\rotatebox[origin=c]{0}{Eagle}} 
  & ED  & $0.0074$ & $0.0058$ & $0.0082$ & $0.0063$ \\
  & ET  & $1.83$   & $1.34$   & $1.90$   & $1.34$   \\
  & \textbf{EF} & $\textbf{98}$      & $146$    & $\textbf{107}$     & $152$   \\
  & EE  & $0.06$   & $0.05$   & $0.07$   & $0.06$   \\
\midrule
\multirow{4}{*}{\rotatebox[origin=c]{0}{Cat}} 
  & ED  & $0.0567$ & $0.0517$ & $0.0730$ & $0.0496$ \\
  & ET  & $1.35$   & $1.56$   & $1.42$   & $1.58$   \\
  & \textbf{EF} & $\textbf{1523}$    & $2070$    & $\textbf{1582}$    & $1813$   \\
  & EE  & $0.31$   & $0.29$   & $0.38$   & $0.28$   \\
\midrule
\multirow{4}{*}{\rotatebox[origin=c]{0}{Deer}} 
  & ED & $0.0201$ & $0.0178$ & $0.0203$ & $0.0211$ \\
  & ET & $1.89$   & $1.22$ & $1.93$   & $1.21$   \\
  & \textbf{EF}   & $\textbf{446}$    & $1086$ & $\textbf{380}$    & $1325$ \\
  & EE  & $0.14$   & $0.13$ & $0.14$   & $0.15$  \\
\midrule
\multirow{4}{*}{\rotatebox[origin=c]{0}{Dog}} 
  & ED & $0.0627$ & $0.0519$ & $0.0636$ & $0.0548$ \\
  & ET & $1.67$   & $1.55$ & $1.70$   & $1.49$   \\
  & \textbf{EF} & $\textbf{772}$    & $949$  & $\textbf{685}$    & $1090$   \\
  & EE  & $0.34$   & $0.29$ & $0.34$   & $0.31$  \\
\botrule
\end{tabular*}
\end{table}

\begin{table}[ht]
\caption{\justifying
EF reduction for Sobel compared to QHED for QPIE and FRQI image encoding methods. The reduction in the number of Edge Fragments (EF), one of the edge-quality metrics used in this study. `Sobel' and `QHED' denote classical Sobel-based post-processing applied after quantum encoding and lag-2 difference computation, the modified quantum Hadamard-based edge detection pipeline, respectively. Overall, under both QPIE and FRQI encodings, classical Sobel-based processing applied to quantum-derived gradients produces more coherent and less noisy edge representations compared to the QHED approach. This observation indicates that the proposed quantum gradient kernel is better suited as a Sobel-like primitive within a hybrid pipeline rather than as a standalone edge detector.
}
\label{tab:ef_reduction}
\centering
\begin{tabular*}{\linewidth}{@{\extracolsep\fill}lcc}
\toprule%
 & \multicolumn{2}{c}{\textbf{EF Reduction (\%)}} \\\cmidrule{2-3}
\textbf{Image} & \textbf{QPIE} & \textbf{FRQI} \\
\midrule
image-1  & 69.07 & 67.20 \\
image-2  & 79.21 & 87.42 \\
image-3  & 22.60 & 47.78 \\
image-4  & 29.27 & 31.79 \\
Eagle    & 32.88 & 29.61 \\
Cat      & 26.42 & 12.74 \\
Deer     & 58.93 & 71.32 \\
Dog      & 18.66 & 37.15 \\
\botrule
\end{tabular*}
\end{table}
\noindent Using the four edge--quality metrics with from Equation~\ref{eqED} to  Equation~\ref{eqEE}, we compared Sobel and QHED under both QPIE and FRQI encodings (tables and qualitative panels). When analyzing Tables~\ref{tab:edge} and ~\ref{tab:ef_reduction}, it is noticeable that, Sobel consistently produces far fewer edge fragments than QHED across all images, with relative EF reductions ranging from roughly $20–-80\%$ ($69–-79\%$ on $image-1/2$ and $59–-71\%$ on Deer), indicating substantially improved edge continuity. This advantage is accompanied by comparable or slightly lower ED, suggesting fewer spurious responses, and by larger average ET, reflecting more coherent ridge-like structures. By contrast, QHED tends to yield lower ET but at the cost of increased fragmentation and, for several Urban100 scenes, higher EE, which is consistent with the visually speckled and over-responsive maps in the overlays. The same trends hold for both encodings—--FRQI shows some of the largest EF gains for Sobel (e.g., image-2), while QPIE achieves similar accuracy with marginally lower randomness on some natural images.\\
Overall, these results consistently indicate that Sobel-based processing produces more coherent and structured edge representations, with lower fragmentation (EF) and generally larger edge thickness (ET), compared to QHED. In contrast, QHED tends to generate more fragmented and noise-sensitive edge maps across both encoding schemes. These findings suggest that quantum gradient computation is more effectively leveraged when integrated within classical operators such as Sobel, rather than being used as a standalone edge detection approach.
Furthermore, classical Canny edge detection results are included in Figures 8 and 9. The Canny detector produces more continuous and smoother edge structures due to its non-maximum suppression and hysteresis thresholding stages, demonstrating strong robustness to noise. In contrast, the proposed quantum-based edge detection methods, particularly QHED, tend to generate more fragmented edge responses while remaining sensitive to high-frequency intensity variations. This comparison further supports the observation that the proposed quantum gradient kernel is more effectively utilized as a Sobel-like primitive within a hybrid framework rather than as a standalone edge detector.
\subsection*{Corner Detection}\label{subsec8}
To quantitatively assess the quality of the detected corners, we use two standard metrics.\\
\noindent
\textbf{Corner Detection Accuracy (CDA):} Measures the ratio of correctly detected corners to the total number of ground-truth corners using Equation ~\ref{eqCDA}.
\begin{equation}
\mathrm{CDA} = \frac{\text{True Positives (TP)}}{\text{Total Ground-Truth Corners}}
\label{eqCDA}
\end{equation}
\vspace{6pt}

\noindent
\textbf{False Positive Rate (FPR):} Measures the proportion of detected corners that do not correspond to any ground-truth corner within a threshold using Equation ~\ref{eqFPR}.
\begin{equation}
\mathrm{FPR} = \frac{\text{False Positives (FP)}}{\text{Total Detected Corners (= TP+FP)}}
\label{eqFPR}
\end{equation}

\noindent A detected corner is counted as a true positive (TP) if it lies within a radius of $r=2$ pixels from any corner of ground--truth; otherwise, it is counted as a false positive (FP).

\noindent\textbf{Classical Harris baseline for corner detection.}
In addition to the Sobel-based and QHED-based quantum pipelines, we include a
\emph{classical Harris corner detector baseline} to evaluate the final corner-detection
quality against a standard non-quantum reference.
Specifically, the classical baseline computes gradients $(I_x,I_y)$ directly in the pixel
domain using the classical Sobel kernels (Equations 32--33), forms the structure tensor
$M$ (Equation 31), and evaluates the Harris response $R(M)$ (Equation 34), followed by
the same non-maximum suppression and thresholding strategy used throughout the study.
This baseline enables a direct comparison between (i) a fully classical Harris detector and
(ii) Harris-style corner extraction driven by quantum-derived lag-2 gradient primitives.

\begin{figure*}[h]
    \centering
    \includegraphics[scale=0.09]{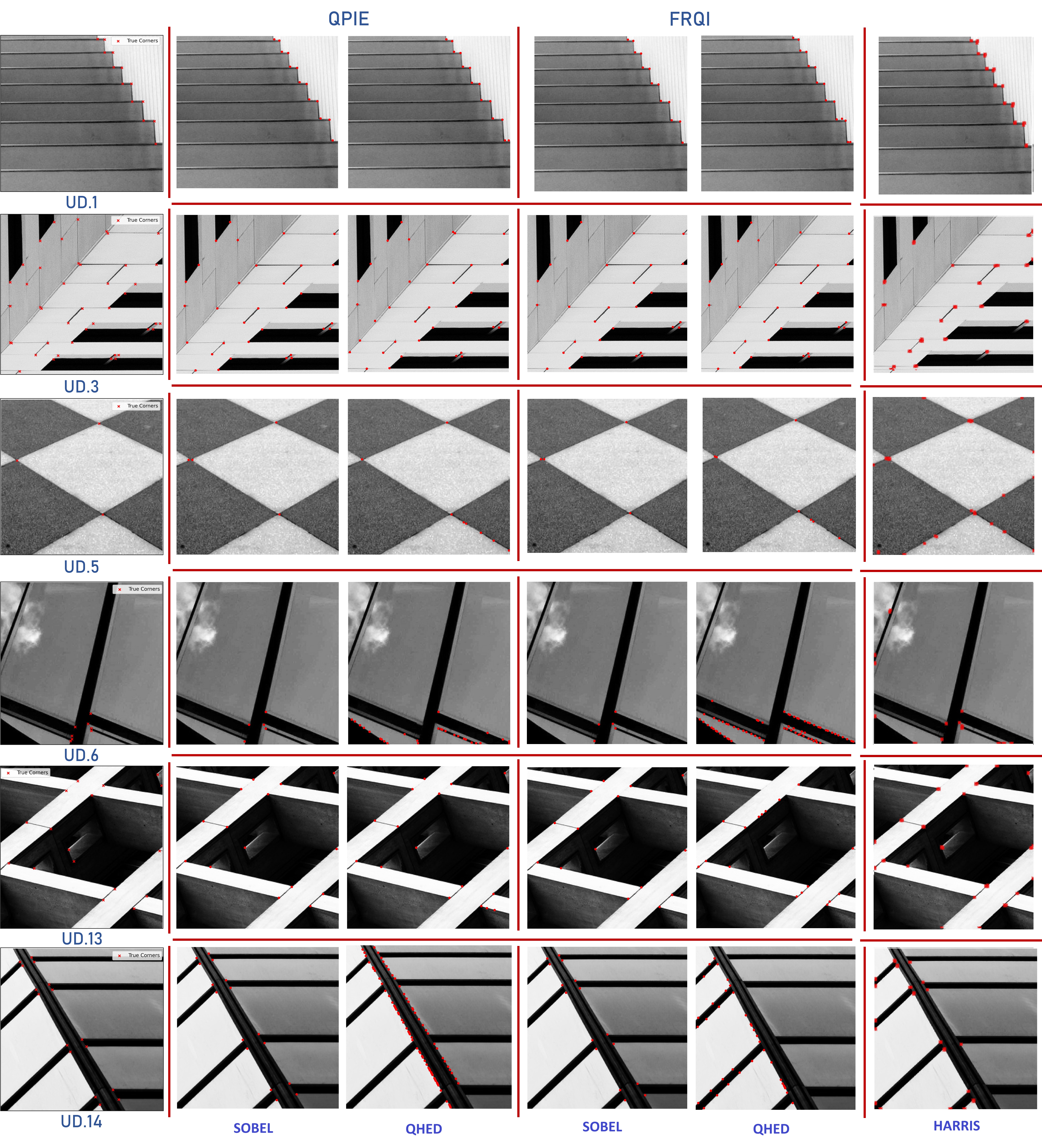}
    \caption{Comparison of QPIE and FRQI corner detection \cite{urban_corner_dataset}. Classical Harris corner detection results are also included for comparison. Harris produces more stable and well-localized corner detections, while QHED-based approaches tend to generate clustered responses along edges, leading to higher false positives.}
    \label{fig:corner_detection}
\end{figure*}

\noindent We evaluate two quantum image encodings (QPIE and FRQI) combined with two corner-extraction pipelines (Sobel and QHED) on representative Urban100 images (UD.1, UD.3, UD.5, UD.6, UD.13, UD.14). The numerical results are summarized in Table~\ref{tab:corner-detection-full}, and qualitative overlays are shown in Figure~\ref{fig:corner_detection}. In Table~\ref{tab:corner-detection-full}, the term `QHED' refers to a corner extraction pipeline derived from QHED-based edge maps rather than a standalone corner detector. In addition, a purely classical Harris corner detector is included as a baseline, where corner detection is performed directly on the original image without quantum encoding, using the same thresholding and non-maximum suppression strategy to ensure a fair comparison. Specifically, the QHED circuit is first applied to generate edge representations using lag-2 Hadamard-based differencing, and corner candidates are subsequently identified using a simple heuristic based on edge intersections and high-response regions. This approach differs from the Sobel-based pipeline, where corner detection is performed using a Harris-style formulation (QHCD) with explicitly computed gradient information and structure tensor analysis. Therefore, the QHED-based method relies on edge-derived structural cues, while the Sobel-based method directly computes corner scores.

\begin{figure*}[h]
    \centering
    \includegraphics[scale=0.09]{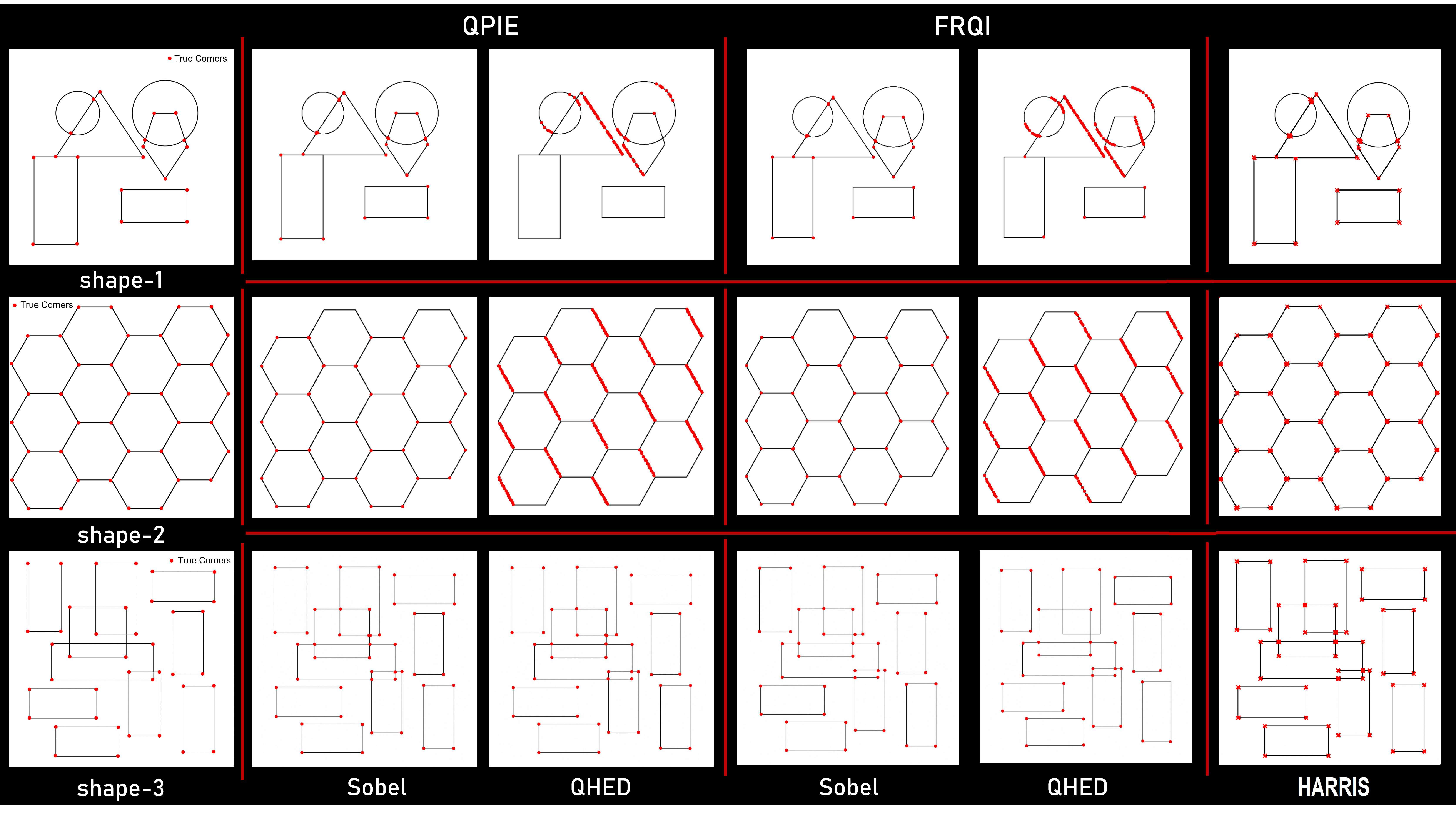}
    \caption{Comparison of QPIE and FRQI corner detection \cite{Huang-CVPR-2015,urban_corner_dataset}Classical Harris corner detection results are also included for comparison. Harris produces more stable and well-localized corner detections, while QHED-based approaches tend to generate clustered responses along edges, leading to higher false positives.}
    \label{fig:corner_detection2}
\end{figure*}

\begin{table}[ht]
\caption{Corner Detection Accuracy (CDA), False Positive Rate (FPR), and corresponding True Positives (TP) and False Positives (FP) for different quantum encoding methods (QPIE, FRQI) and corner detection pipelines across selected images from the Urban100 dataset. Here, `Sobel' refers to Harris-style corner detection (QHCD) using Sobel-based gradients derived from the quantum lag-2 primitive. In contrast, `QHED' refers to a corner detection pipeline derived from QHED-based edge maps obtained via the lag-2 Hadamard-based edge detection method, where corner candidates are extracted using an edge-based heuristic rather than a direct Harris response computation. Furthermore, `Harris' denotes the classical Harris corner detector applied directly to the original image without quantum encoding, using the same thresholding and non-maximum suppression strategy for fair comparison. The images are taken from the Urban100 dataset~\cite{Huang-CVPR-2015,urban_corner_dataset}.}
\label{tab:corner-detection-full}
\centering
\begin{tabular*}{0.95\linewidth}{@{\extracolsep\fill}lcccccc}
\toprule
\textbf{\# True Corners} & \textbf{Encoding}& \textbf{Method} & \textbf{TP} & \textbf{FP} & \textbf{CDA (\%)} & \textbf{FPR (\%)} \\
\midrule
\multirow{5}{*}{UD.1 - 13}
  & \multirow{2}{*}{QPIE} & \textbf{Sobel} & \textbf{13} & \textbf{0} & \textbf{100.00} & \textbf{0.00} \\
  &  & QHED & 13 & 1 & 100.00 & 7.14 \\
  &  \multirow{2}{*}{FRQI} & \textbf{Sobel} & \textbf{13} & \textbf{0} & \textbf{100.00} & \textbf{0.00} \\
  &  & QHED & 13 & 2 & 100.00 & 13.33 \\
  &  & \textbf{Harris} & \textbf{13} & \textbf{0} & \textbf{100.00} & \textbf{0.0} \\
\midrule
\multirow{5}{*}{UD.3 - 43}
  & \multirow{2}{*}{QPIE} & \textbf{Sobel} & \textbf{32} & \textbf{0} & \textbf{74.42} & \textbf{0.00} \\
  &  & QHED & 33 & 2 & 76.74 & 5.71 \\
  &  \multirow{2}{*}{FRQI} & \textbf{Sobel} & \textbf{29} & \textbf{0} & \textbf{67.44} & \textbf{0.00} \\
  &  & QHED & 30 & 0 & 69.77 & 0.00 \\
  &  & \textbf{Harris} & \textbf{18} & \textbf{1} & \textbf{41.86} & \textbf{5.26} \\
\midrule
\multirow{5}{*}{UD.5 - 3}
  & \multirow{2}{*}{QPIE} & \textbf{Sobel} & \textbf{3} & \textbf{1} & \textbf{100.00} & \textbf{25.00} \\
  & & QHED & 3 & 5 & 100.00 & 62.50 \\
  &  \multirow{2}{*}{FRQI} & \textbf{Sobel} & \textbf{3} & \textbf{1} & \textbf{100.00} & \textbf{25.00} \\
  &  & QHED & 3 & 4 & 100.00 & 57.14 \\
  &  & \textbf{Harris} & \textbf{3} & \textbf{17} & \textbf{100.00} & \textbf{85.00} \\
\midrule
\multirow{5}{*}{UD.6 - 8}
  & \multirow{2}{*}{QPIE} & \textbf{Sobel} & \textbf{5} & \textbf{0} & \textbf{62.50} & \textbf{0.00} \\
  & & QHED & 3 & 25 & 37.50 & 89.29 \\
  & \multirow{2}{*}{FRQI} & \textbf{Sobel} & \textbf{5} & \textbf{0} & \textbf{62.50} & \textbf{0.00} \\
  & & QHED & 3 & 65 & 37.50 & 95.59 \\
  &  & \textbf{Harris} & \textbf{5} & \textbf{5} & \textbf{62.5} & \textbf{50.0}\\
\midrule
\multirow{5}{*}{UD.13 - 17}
  & \multirow{2}{*}{QPIE} & \textbf{Sobel} & \textbf{15} & \textbf{0} & \textbf{88.24} & \textbf{0.00} \\
  & & QHED & 15 & 2 & 88.24 & 11.76 \\
  & \multirow{2}{*}{FRQI} & \textbf{Sobel} & \textbf{15} & \textbf{0} & \textbf{88.24} & \textbf{0.00} \\
  & & QHED & 10 & 7 & 58.82 & 41.18 \\
  &  & \textbf{Harris} & \textbf{14} & \textbf{4} & \textbf{82.35} & \textbf{22.22} \\
\midrule
\multirow{5}{*}{UD.14 - 16}
  & \multirow{2}{*}{QPIE} & \textbf{Sobel} & \textbf{16} & \textbf{0} & \textbf{100.00} & \textbf{0.00} \\
  & & QHED & 15 & 78 & 93.75 & 83.87 \\
  & \multirow{2}{*}{FRQI} & \textbf{Sobel} & \textbf{16} & \textbf{0} & \textbf{100.00} & \textbf{0.00} \\
  & & QHED & 8 & 29 & 50.00 & 78.38 \\
  &  & \textbf{Harris} & \textbf{13} & \textbf{5} & \textbf{81.25} & \textbf{27.77} \\
\botrule
\end{tabular*}
\end{table}

\noindent The quantitative results in Table~\ref{tab:corner-detection-full} show that the QHCD attains high CDA with very low FPR across most images under both QPIE and FRQI encodings, yielding clean and well-localized responses. In more textured or repetitive scenes such as UD.3 and UD.6, QHED frequently fires at multiple points along single edge segments, which inflates FPR—often to 90\% (and up to 96\% for FRQI)—despite comparable CDA. The qualitative overlays in Figure~\ref{fig:corner_detection} and \ref{fig:corner_detection2} corroborate these trends, with Sobel producing more isolated, coherent corner detections and QHED exhibiting clustered, spurious responses. Eventually, Table~\ref{tab:corner-detection-full}, Figure~\ref{fig:corner_detection} and \ref{fig:corner_detection2} indicates that the Sobel-based quantum pipeline is superior to QHED for reliable corner extraction with both QPIE and FRQI encoding models.
In addition, classical Harris corner detection results are included in Figures 10 and 11 to provide a direct baseline comparison. The classical Harris detector produces more stable and well-localized corner points, while the quantum-derived methods may exhibit denser or more scattered detections depending on the underlying gradient representation. In particular, QHED-based corner extraction tends to generate clustered responses along edges, leading to increased false positives, whereas Sobel-based quantum gradients yield more isolated and structurally meaningful corner detections. These observations highlight the importance of integrating quantum-derived gradients within classical corner detection frameworks for improved reliability.
\noindent Broader contextualization against classical state-of-the-art detectors.
Our experimental comparisons focus on isolating the effect of the \emph{quantum gradient primitive} (lag-2 differencing) by benchmarking against the classical Sobel operator applied within the same post-processing pipeline. To situate detection quality more broadly, it is important to note that state-of-the-art classical edge/corner extraction often relies on multi-stage or learned detectors, e.g., Canny edge detection with non-maximum suppression and hysteresis thresholding, or modern learned interest-point detectors that jointly predict
keypoint confidence and descriptors. These methods can offer improved robustness to noise, texture, and illumination changes relative to simple gradient-thresholding pipelines.

From an accuracy-scaling perspective, the quantum kernel computes gradient information coherently in amplitude space after a single circuit execution, but any conversion to dense classical maps is ultimately limited by measurement statistics. As shot count increases, the estimated measurement outcomes converge to the underlying ideal probabilities, and thus the derived gradient-based scores converge to their classical counterparts. Therefore, the accuracy of quantum-derived gradients scales with the measurement budget, while the circuit depth required to \emph{generate} the gradient-encoded state remains polylogarithmic
in the number of pixels. A full empirical comparison against advanced classical detectors (e.g., Canny and learned keypoint detectors) constitutes an important extension of this work and will be explored in future experiments for comprehensive positioning.\\
\noindent\textbf{Computational complexity considerations.} In addition to detection quality, it is important to discuss the computational cost of the proposed quantum circuits. The key subroutines in our approach, including the cyclic shift (permutation) operator and lag-2 differencing, operate on the position register and can be implemented using $O(r)$ quantum gates, where $r$ denotes the number of position qubits and $N = 2^r$ is the number of encoded pixel positions. In particular, the cyclic shift corresponds to an addition modulo $2^r$, which can be efficiently realized using standard quantum adder circuits with linear gate complexity.
In contrast, classical Sobel-based edge and corner detection on an image of size $512 \times 512$ requires $O(N)$ pixel-wise convolution operations, where $N = 512^2$. Therefore, while classical methods scale linearly with the number of pixels, the proposed quantum subroutines scale polynomially with the number of qubits (i.e., logarithmically in the number of pixels).
However, it should be emphasized that the present implementation is based on statevector simulation and does not account for state preparation, measurement overhead, or hardware noise. As a result, the current work demonstrates a potential subroutine-level computational advantage rather than a full end-to-end performance gain. A practical advantage would require large-scale fault-tolerant quantum hardware, which remains an important direction for future work.
\section{Conclusion}\label{sec4}
In this paper, we have implemented the Quantum Harris Corner Detection (QHCD) algorithm using quantum computing technologies. Our approach has demonstrated more reliable performance within the proposed framework compared to the Quantum Hadamard Edge Detection (QHED) method, particularly in terms of accurately detecting structural discontinuities such as corners. Furthermore, experimental results indicate that the QHCD algorithm, when integrated with the QPIE model, shows more stable and structurally coherent results compared to its implementation with the FRQI  model. This improvement can be attributed to the amplitude-based representation used in QPIE, which offers more stable and compact encoding for spatial variations. By contrast, the angle-based (sin/cos) parameterization in FRQI tends to amplify small hyperparameter perturbations, complicating robust corner localization.\\
We have also observed that the accuracy of the QHCD algorithm is sensitive to certain design parameters, such as the selected hyperparameter values and the image resolution (i.e., number of pixels encoded). Increasing the number of pixels and the corresponding number of qubits enhances spatial granularity and yields more reliable corner detection results. However, higher resolution also increases state-preparation overhead and circuit depth, which must be balanced against shot budgets and hardware noise on near-term devices.It should be noted that all experiments in this study are conducted under noiseless simulation settings, and practical performance on NISQ hardware may be affected by noise, measurement overhead, and finite-shot statistics.\\
In future work, the method and the algorithm we have proposed for detecting corners can be improved by systematic tuning of hyperparameters within the QHCD framework. Another possibility to improve our presented results may be increasing the number of pixels of the images and qubit numbers. Furthermore, the proposed approach provides a flexible and promising framework for quantum corner detection, which can be integrated with broader quantum image processing and machine learning pipelines, such as the classification of images depending on the corner number involving quantum image processing and machine learning. In addition, data-driven hyperparameter tuning (e.g., Bayesian or gradient-based search) could stabilize performance across datasets and back-ends without substantially increasing circuit depth.

\noindent FRQI relies on angle-based (sin/cos) encoding, which introduces higher sensitivity to hyperparameter selection, particularly in thresholding operations within the Harris response. This sensitivity can lead to less stable corner detection compared to the amplitude-based QPIE encoding, which exhibits more robust behavior.

\section*{Acknowledgements}

This work was initiated as part of the QIntern 2024 program.

\section*{Competing Interests}
\noindent No funding was received for conducting this study.

\section*{Declarations}
\noindent \textbf{{\large Data availability}} 

\noindent All data generated or analyzed during this study are included in this article.

\noindent \textbf{{\large Code availability}}

\noindent All the experimental results and source code implementations are available at \href{https://github.com/mdaamirQ/QHCD}{https://github.com/mdaamirQ/QHCD}.

\section*{Authors Contribution}
\textbf{Simulations of QPIE edge detection, comparison:} Simge and Asude work with Mohammad.\\
\textbf{FRQI edge detection algorithm and comparison:} Batuhan Hangun Emre worked with Gabriela. \\
\textbf{Problem formulation:} Yasemin and Simge. \\
\textbf{Supervision of the project:} Yasemin gave us the tutorial and resources for understanding quantum image processing. \\
\textbf{Detection of edge and corner with QPIE and FRQI:} Gabriela and Mohammad. \\
\textbf{Paper writing:} Yasemin wrote the introduction and related works. Mohammad wrote the technical part of the QHCD. Gabriela and Mohammad completed the experimental results.\\

\bibliography{sn-bibliography}

\end{document}